\pdfoutput=1
\documentclass[11pt]{article}

\usepackage{EMNLP2023}

\usepackage{times}
\usepackage{latexsym}
\usepackage{xspace,mfirstuc,tabulary}
\usepackage{inconsolata}
\usepackage[misc]{ifsym}
\usepackage{xcolor}
\usepackage{graphicx}
\usepackage{multirow}
\usepackage{booktabs}
\usepackage{amsmath,amsfonts,amssymb}
\usepackage{hyperref}
\usepackage{pifont}
\usepackage{enumitem}
\usepackage{caption}
\usepackage{amsfonts}
\usepackage{array, makecell}
\usepackage{subcaption}
\usepackage{tikz}

\usepackage[T1]{fontenc}
\usepackage[utf8]{inputenc}

\usepackage{microtype}

\usepackage{inconsolata}

\title{Auto-Instruct: Automatic Instruction Generation and Ranking \\ for Black-Box Language Models}
\author{{\bf Zhihan Zhang\textsuperscript{\Letter}$^{\spadesuit}$\thanks{\hspace{0.175cm}This work was done when Zhihan was an intern at Microsoft Azure AI.}\hspace{0.1cm}, Shuohang Wang$^{\diamondsuit}$, Wenhao Yu$^{\spadesuit}$, Yichong Xu$^{\diamondsuit}$, Dan Iter$^{\diamondsuit}$},\\ {\bf Qingkai Zeng$^{\spadesuit}$, Yang Liu$^{\diamondsuit}$, Chenguang Zhu$^{\diamondsuit}$, Meng Jiang$^{\spadesuit}$} \\
$^{\spadesuit}$University of Notre Dame \\
$^{\diamondsuit}$Microsoft Azure AI \\
\normalsize{ {\tt zzhang23@nd.edu}}
}

\begin{document}
\maketitle
\begin{abstract}
% \vspace{-0.1cm}
% In this paper, we introduce Auto-Instruct, an innovative and effective approach for the automated optimization of instructions for large language models (LLMs). Natural language instructions play a crucial role in directing instruction-tuned black-box LLMs, yet manual instruction engineering remains the prevailing method in many NLP tasks. While there have been attempts to prompt LLMs to generate instructions autonomously, these often rely on validating instructions against additional validation sets, a resource not available in the true few-shot setting. Our approach begins by prompting the LLM to generate a variety of candidate instructions in diverse styles, before ranking these instructions with a model trained on 575 tasks. In experiments on 118 out-of-domain tasks, our approach has shown remarkable performance, surpassing both human-written instructions and existing LLM-prompting methods.

Large language models (LLMs) can perform a wide range of tasks by following natural language instructions, without the necessity of task-specific fine-tuning. Unfortunately, the performance of LLMs is greatly influenced by the quality of these instructions, and manually writing effective instructions for each task is a laborious and subjective process. In this paper, we introduce Auto-Instruct, a novel method to automatically improve the quality of instructions provided to LLMs. Our method leverages the inherent generative ability of LLMs to produce diverse candidate instructions for a given task, and then ranks them using a scoring model trained on a variety of 575 existing NLP tasks. In experiments on 118 out-of-domain tasks, Auto-Instruct surpasses both human-written instructions and existing baselines of LLM-generated instructions. Furthermore, our method exhibits notable generalizability even with other LLMs that are not incorporated into its training process.\footnote{Model and code are available at \url{https://github.com/ytyz1307zzh/Auto-Instruct}.}

\end{abstract}

\section{Introduction}

Instruction-tuned large language models (LLMs) have gained considerable popularity as solutions to a myriad of NLP tasks, owing to their proficiency in interpreting natural language instructions~\cite{flan, flan-t5, InstructGPT, alpaca}. As fine-tuning LLMs often becomes unfeasible, instructions play an increasingly crucial role in prompting such black-box LLMs. Especially in the \textit{true few-shot}
% \footnote{The term \textit{true few-shot}, as first introduced in ~\citet{true_few_shot}, denotes a learning scenario with no available held-out examples (i.e., validation set) to fine-tune various aspects of the learning process. These include aspects like hyperparameters, training objectives, instructions, prompts, and etc.} 
\footnote{A scenario where no additional training or validation data are available for hyperparameter tuning and prompt selection, in addition to the few-shot examples~\cite{true_few_shot}.}
setting~\cite{true_few_shot} where the user aims to tackle a new task with only a basic task description and a few data examples at hand, a well-crafted instruction is imperative in enabling the LLM to grasp the required input-output mapping to complete the task.

% Instruction-tuned language models (LMs) have been popular NLP solutions
% % gained increasingly popularity in NLP recently
% ~\cite{flan, flan-t5, InstructGPT, alpaca}. These LMs are fine-tuned to understand and follow natural language instructions, enabling them to handle a wide range of tasks with a single model, provided task-specific instructions.
% % are given.
% An instruction, in this context, refers to a textual description of the task, encompassing the explanation of the input, constraints on the output format, or guidance on generating the expected output based on the given input. With the growing size of large LMs and the associated cost of fine-tuning them, many powerful LMs are encapsulated as black-box models which are only accessible via API
% % interfaces
% ~\cite{InstructGPT, gpt-4}, emphasizing the importance of instructions. As fine-tuning large LMs using downstream data becomes impractical, instructions
% themselves
% have become a critical source for these models to comprehend the expected input-output mapping required to accomplish a given task.

\begin{figure}[t]
    \includegraphics[width=0.5\textwidth]{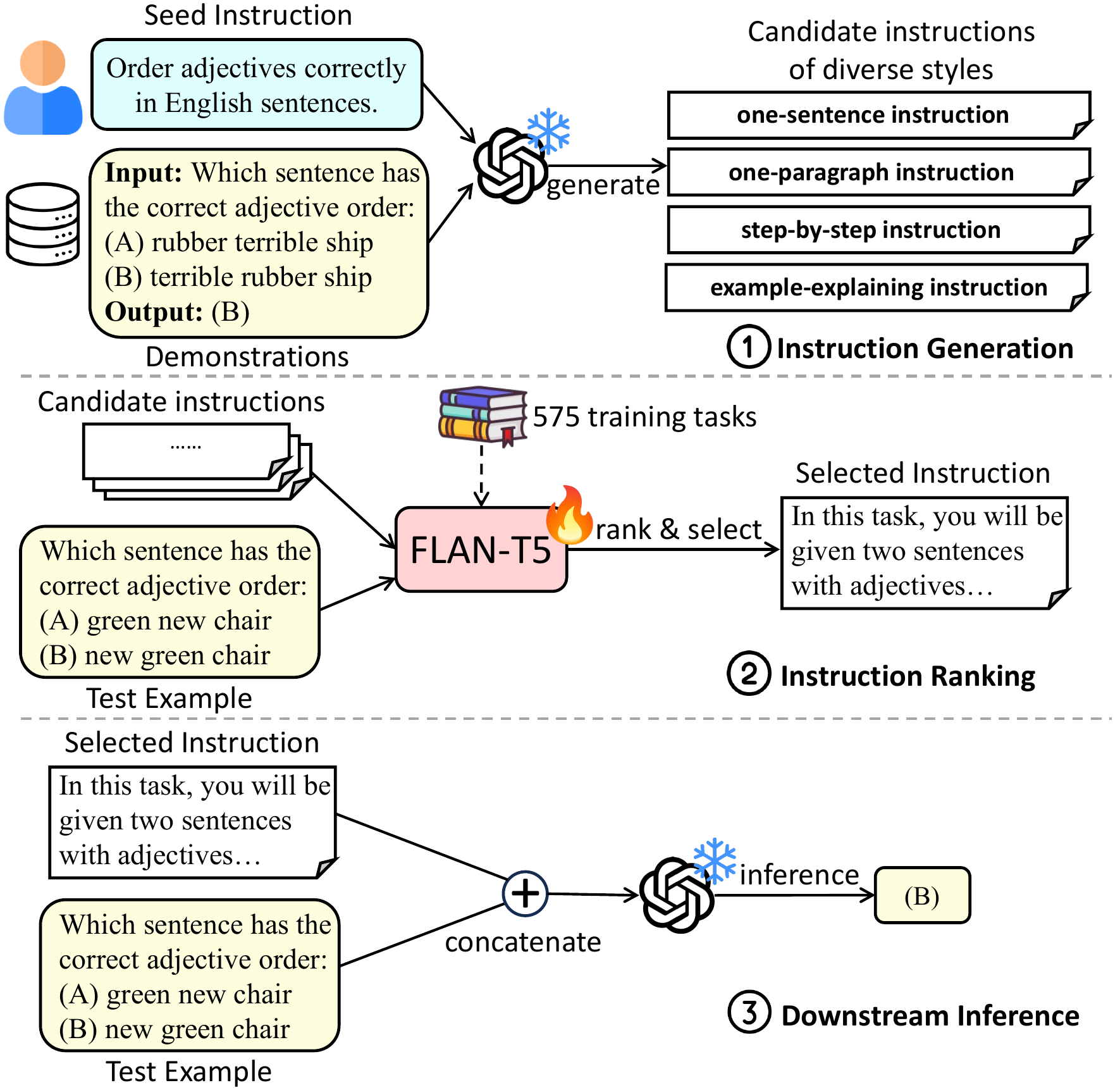}
    \vspace{-0.5cm}
  \caption{The Auto-Instruct pipeline. We first prompt the LLM to generate a diverse set of candidate instructions with different styles, and then train a model to rank and select the most effective instruction for a given example. Finally, the selected instruction is used to prompt the LLM to infer the output for this example.}
  \vspace{-0.2cm}
  \label{fig:introduction}
\end{figure}

Despite the significance of instructions, the prevailing approach when using a black-box LLM on a new task remains to be manual 
prompt engineering~\cite{chatgpt_prompt_catalog, promptaid}. Such an approach, however, is not only time-consuming but also tends to yield suboptimal instructions. Against this backdrop, efforts have been made to empower LLMs to generate instructions automatically~\cite{instruction_induction, human_level_instruction_engineer, iprompt}. These approaches feed the LLM a handful of examples and prompt it to generate an instruction based on these demonstrations. While such methods showcase the LLM's capability to generate coherent instructions~\cite{instruction_induction}, only generating a single instruction cannot guarantee reliable performance for unseen examples in the given task. As a straightforward solution, validation sets have been used to evaluate the effectiveness of a set of sampled instructions~\cite{human_level_instruction_engineer, iprompt}, but this is impracticable for many tasks defined under the true few-shot setting~\cite{bbh}. Besides, these approaches have primarily been tested on simple tasks where basic instructions are already sufficient, such as arithmetic operations or sentiment classification. More complex tasks in NLP benchmarks~\cite{superNI}, which necessitate careful instruction engineering, remain largely unexamined for an automatic solution.

To address the aforementioned challenges, we propose Auto-Instruct, a novel approach to automatically generate and rank instructions for black-box LLMs across various NLP tasks, under the true few-shot setting. For each downstream task, we first prompt the LLM to sample a variety of candidate instructions, based on a basic seed instruction and few-shot demonstrations. We collect a diverse candidate set by specifying the expected \textit{style} of each instruction. Recognizing the variable performance of LLMs across different instructions, coupled with the lack of validation data for pre-emptive instruction selection, we train a scoring model to rank and select the most appropriate instruction for each downstream test example. To ensure necessary generalizability in the few-shot setting, the model is trained on 575 exisiting NLP tasks before being deployed for out-of-domain test tasks. Finally, the selected instruction is used to prompt the LLM for downstream inference.

In experiments with OpenAI's \textit{text-davinci-003}, Auto-Instruct yields remarkable performance on 118 out-of-domain tasks from Super Natural Instructions (SuperNI;~\citealp{superNI}) and Big Bench Hard (BBH;~\citealp{bbh}). Showing robust generalizability in out-of-domain scenarios, Auto-Instruct outperforms human-written seed instructions, the state-of-the-art instruction generation approach iPrompt~\cite{iprompt}, and various baselines of prompting the LLM for instruction selection. Moreover, Auto-Instruct exhibits impressive performance in the zero-shot setting and in generalization to other LLMs (\textit{i.e.}, ChatGPT and GPT-4). Our study underlines that automatically generating and ranking instructions is a promising approach for leveraging the power of black-box LLMs effectively.

\section{Related Work}

% Due to the importance of instruction in using LLMs, various approaches have been made to optimize the instruction. The first line of work focuses on the \textit{parametric optimization} of instructions. 
The choice of instructions plays a pivotal role in effectively utilizing LLMs. To this end, a range of approaches has been implemented, with \textit{parametric optimization} and \textit{LLM-based generation} standing out as prominent methods. Parametric optimization primarily involves utilizing parameters to tune instructions~\cite{autoprompt,fluentprompt,RLprompt}. For instance, \citet{autoprompt} employed a gradient-based search over a predetermined length of discrete tokens as the instruction. \citet{fluentprompt} further improved this approach by preserving the readability of the sampled tokens through a perplexity constraint. As a more flexible approach, \citet{RLprompt} optimized instruction generation through reinforcement learning, with rewards computed based on the LLM output. However, these strategies require access to either LLM parameters or a training set for optimization, making them less applicable to black-box LLMs with only a limited number of available examples. Moreover, instructions generated by these methods often lack fluency or even become gibberish, thereby compromising their interpretability.

% The other line of work, meanwhile, attempted \textit{non-parametric optimization} of instructions by purely prompting the LLM. 
In contrast, the LLM-based generation thread selects instructions by directly prompting the LLM~\cite{instruction_induction,human_level_instruction_engineer,iprompt}.
For example, \citet{instruction_induction} were among the first to reveal that LLMs could write an instruction for a given task after observing just a few demonstrations, and \citet{human_level_instruction_engineer} improved the quality of the generated instructions by selecting the best performed one on the validation data. iPrompt~\cite{iprompt} is the most capable method so far with its iterative generation and validation process for selecting instructions. Nevertheless, these approaches still necessitate a validation set for instruction ranking, and the instructions they generate typically underperform compared to those written by humans.

Besides the choice of instructions, researchers have also explored other orthogonal directions of improving LLM prompts, such as the selection of in-context demonstrations. Some works focused on identifying the most suitable demonstrations from training examples~\cite{demo_retrieval, dynaic_prompt_engineering, LLM_are_topic_models} and their optimal ordering~\cite{demo_order} in the few-shot prompt. Other studies examined the engineering and selection of reasoning chains that are paired with the few-shot demonstrations on multi-step reasoning tasks~\cite{CoT, auto_cot, select_cot, liang2023mint}. We reserve the exploration of integrating these orthogonal techniques with our approach to holistically optimize the entire LLM prompt for future work.

\section{Problem Formulation}
\label{sec:formulation}
In this work, we focus on the true few-shot setting where a user aims to tackle a new task with a black-box LLM. While it is easy to come up with a handful of examples and a basic description, the user may not have insights into what kind of instructions would be effective for unseen examples. Hence, given the few-shot examples as demonstrations and the basic description as a seed instruction, our goal is to automate the process of creating a more effective instruction for the given task.

We formulate our problem following the conventional practices of in-context learning~\cite{survey_ICL}. In the aforementioned \textbf{few-shot} setting, the \textit{prompt} to query a black-box LLM comprises an \textit{instruction} $I$, the test input $x$, and a few input-output pairs as \textit{demonstrations} $\{x^d_i, y^d_i\}^n_{i=1}$. The LLM is expected to generate an output $y\sim P(\cdot|I, \{x^d_i, y^d_i\}^n_{i=1}, x)$. This work aims to automatically find a superior instruction $I'$ based on the human-written seed instruction $I^s$, thereby circumventing the need for substantial manual engineering. Besides, we also explore the \textbf{zero-shot} setting where no demonstrations are given to the LLM.

Despite the instruction potentially having multiple ways of integrating with the demonstrations and the test input, to reduce the complexity of the problem, we format the whole prompt in the order of $(I,x^d_1,y^d_1,\cdots,x^d_n,y^d_n,x)$. This aligns with the convention of problem-solving where the task is first outlined, followed by the provision of data examples, and the test input is finally provided. In practice, we use $n=3$ for all tasks.

\section{Auto-Instruct}

% The common practice of describing a task at hand is to manually write an instruction. However, users may face challenges in finding a satisfactory instruction due to the randomness of the manually written instruction and the instruction-dependent performance fluctuations of LLMs. To ease the burden of instruction engineering, attempts have been made to prompt the LLM itself to generate meaningful instructions~\cite{human_level_instruction_engineer, iprompt}. 

Auto-Instruct is composed of two steps: instruction generation and instruction ranking. We first prompt the black-box LLM to generate a diverse set of candidate instructions (\S\ref{sec:generation}) for each downstream task. Next, we train a scoring model to rank all candidate instructions for each given test example, as different examples can benefit from different instructions (\S\ref{sec:ranking}). Then, the top-ranked instruction is selected to prompt the black-box LLM on that specific test example for downstream inference.

\begin{figure}
    \centering\includegraphics[width=0.35\textwidth]{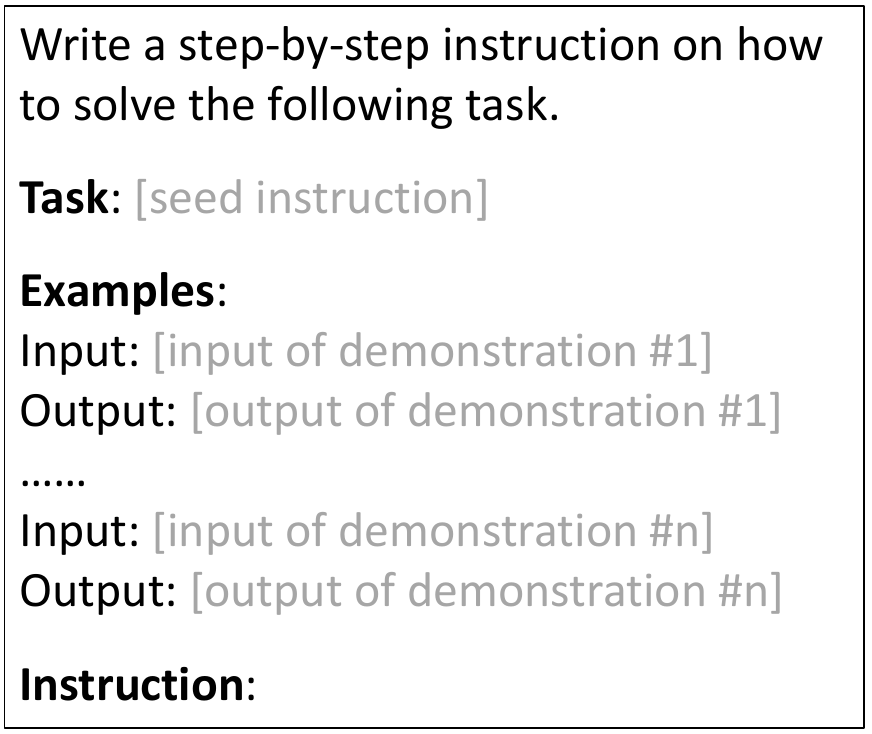}
    \caption{The meta-prompt that guides the LLM to generate a step-by-step instruction for the given task. Other meta-prompts are shown in Appendix~\ref{app:meta_prompt}.}
    \label{fig:meta_prompt}
\end{figure}

\subsection{Instruction Generation}
\label{sec:generation}
As mentioned in \S\ref{sec:formulation}, we leverage a basic human-written task description as the seed instruction $I^s$ and prompt the black-box LLM to generate a number of candidate instructions $\{I^c_j\}^m_{j=1}$. Specifically, in the few-shot setting, we prompt the LLM to generate candidate instructions ${I^c\sim P(\cdot|I^s,\{x^d_i, y^d_i\}^n_{i=1})}$ based on the seed instruction and few-shot demonstrations. Previous approaches~\cite{human_level_instruction_engineer, iprompt} only utilized a single meta-prompt\footnote{The prompt for the LLM to generate instructions.} and collected candidate instructions via token sampling. Usually, such sampled instructions only show minor variations in phrasing rather than substantial content diversity. Moreover, their quality recursively rely on the arbitrary choice of the meta-prompt, which transfers the unreliability of manual instruction engineering to manual meta-prompt engineering. 

In our improved approach, we curate a set of meta-prompts to stimulate the LLM to sample diverse candidate instructions by defining different required \textit{styles} of the instruction. These meta-prompts include:
\begin{enumerate}[noitemsep,topsep=1pt,parsep=1pt,partopsep=0pt]
    \item Write an instruction on how to solve the following task in \textit{one sentence}.
    \item Write an instruction on how to solve the following task in \textit{one paragraph}.
    \item Write a \textit{step-by-step} instruction on how to solve the following task.
    \item Write an instruction on how to solve the following task. The instruction must include the \textit{explanations of the given examples}.
\end{enumerate}
Alongside these 4 meta-prompts, we also bring in human-written instructions from existing NLP tasks as demonstrations to guide the generation of instructions. Intuitively, we prompt the LLM to emulate the style of human-written instructions in these demonstration tasks. We source demonstration tasks with their instructions from our training tasks in SuperNI, grouping them into 3 clusters based on the length of their instructions, so as to guide the LLM to generate instructions of different granularities. Figure~\ref{fig:meta_prompt} provides an example of the meta-prompt \#3. Other meta-prompts are detailed in Appendix~\ref{app:meta_prompt}.

% Besides this list, we also leverage human-written instructions from existing tasks as demonstrations for instruction generation. We sourced these tasks and instructions from our training tasks in SuperNI, grouping them into 3 clusters with different average instruction lengths. By prompting the LLM with sampled instructions from each cluster, the LLM is guided to generate instructions of different granularities. Figure~\ref{fig:meta_prompt} provides an example of the meta-prompt for generating step-by-step instructions. Other meta-prompts are detailed in Appendix~\ref{app:meta_prompt}.

Based on these 7 distinct meta-prompts (\textit{i.e.}, 4 style-specific meta-prompts + 3 groups of demonstration tasks), we sample 3 instructions under each meta-prompt via nucleus sampling~\cite{nucleus}. Including the original seed instruction, we collect a total of 22 candidate instructions for each task. As a result, we create a diverse and comprehensive set of candidate instructions, thereby reducing the randomness brought by the nuances of different meta-prompts.

% By sampling multiple instructions under each meta-prompt, we collect a diverse and comprehensive set of candidate instructions, thereby reducing the randomness brought by the nuances of different meta-prompts. 

In the zero-shot setting, due to the absence of demonstrations, the LLM is prompted to generate the candidate instruction $I^c\sim P(\cdot|I^s)$ based solely on the seed instruction. Besides, the example-explaining meta-prompt is removed. As we demonstrate in \S\ref{sec:case_study}, even without the aid of demonstrations, our style-specific meta-prompts still enable the LLM to generate informative instructions.

\subsubsection{Instability Under Different Instructions}

\begin{figure}
    \centering
    \includegraphics[width=0.48\textwidth]{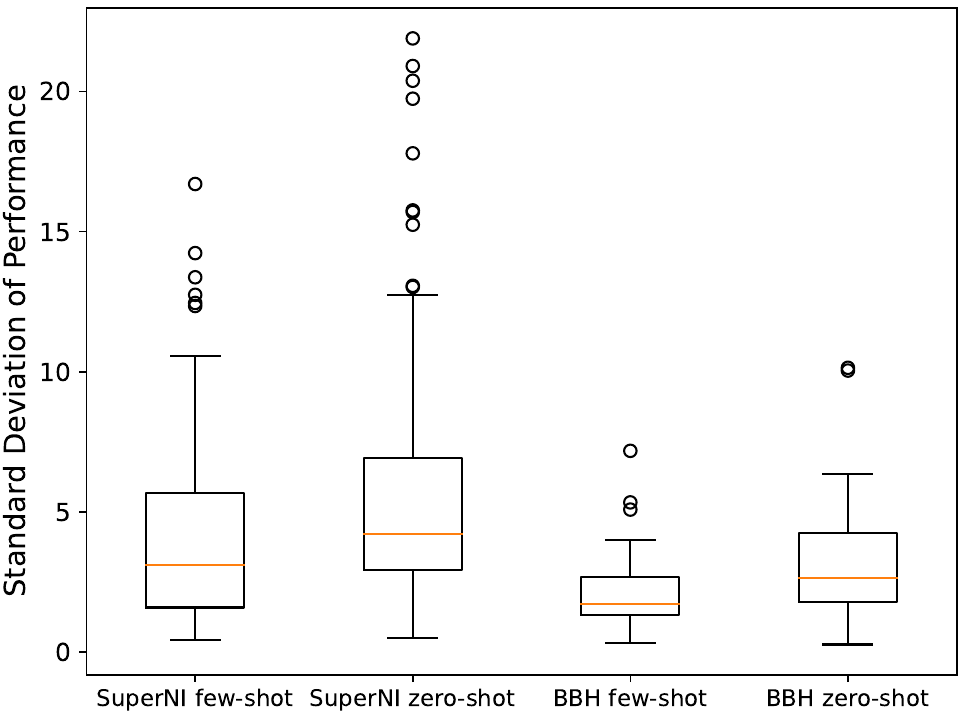}
    \caption{Box plot showing how much the LLM performance varies with different instructions, tested on OpenAI's \textit{text-davinci-003}. Performance is evaluated by ROUGE-L on SuperNI and Accuracy on BBH. Each value represents the standard deviation of LLM performance across all generated instructions on a single task.}
    \label{fig:instruction_stdev}
\end{figure}

While LLMs are capable of generating meaningful instructions, relying on a single generated instruction will probably lead to suboptimal performance due to the LLM's sensitivity to the phrasing of the instructions. This instability is particularly evident in the zero-shot setting due to the lack of demonstrations to assist prediction.
In Figure~\ref{fig:instruction_stdev}, we calculate the standard deviation of LLM performance using different instructions, after having evaluated all instructions for each downstream task. This indicates the expected performance fluctuation when substituting one instruction for another. The median standard deviation across all tasks are 3.1 and 4.2 points in ROUGE-L for few-shot and zero-shot settings respectively on SuperNI, and the upper quartiles are 5.7 and 6.9 points respectively. The choice of instruction even causes double-digit performance fluctuation on many tasks. Therefore, the development of a method to rank and select instructions becomes an essential undertaking.

% As Figure~\ref{fig:instruction_stdev} illustrates, after evaluating all generated instructions on downstream test tasks, the median standard deviation of model performance across all tasks are 3.1 and 4.2 points for few-shot and zero-shot settings respectively on SuperNI, and the upper quartiles are 5.7 and 6.9 points respectively. The choice of instruction even causes double-digit performance fluctuation on many tasks. Consequently, the development of a method to rank and select instructions becomes an essential undertaking.

\subsection{Instruction Ranking}
\label{sec:ranking}

In a true few-shot setting, demonstrations are inadequate to reflect the effectiveness of candidate instructions due to the small sample size. To circumvent this limitation, we train a generalizable instruction ranking model across a variety of NLP tasks, and subsequently apply it to each test example in out-of-domain tasks. Intuitively, this model is trained to rank instructions against their downstream performance on the LLM, \textit{i.e.}, to assign higher scores to more effective instructions. 
% Finally, the model is used to select an appropriate instruction for each test example.

\subsubsection{Model}
\label{sec:model}

Owing to the proven generalizibility of the T5 model family~\cite{T5, t0}, we start from the instruction-tuned FLAN-T5-Large model~\cite{flan-t5} and train it with our instruction ranking objective. Given a specific example $(x,y)$ where $x$ is the input and $y$ is the ground-truth output, as well as an arbitrary candidate instruction $I^c$, the model predicts a score $Q_{\textrm{T5}}(I^c,x)$ as an estimate of the instruction's effectiveness on the example. Leveraging the instruction-following nature of FLAN-T5, we give the following prompt to the ranking model:

\vspace{-0.2cm}
\begin{figure}[h!]
    \centering
    \includegraphics[width=0.42\textwidth]{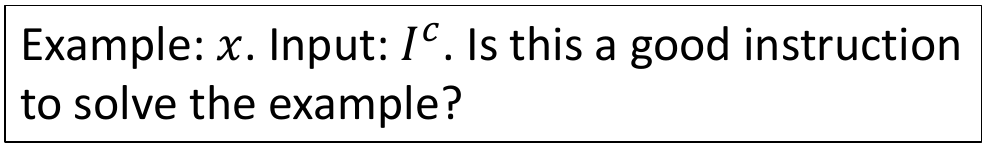}
\end{figure}
\vspace{-0.3cm}
\noindent $Q_{\textrm{T5}}(I^c,x)$ is then calculated as the logit of the ``yes'' token at the starting position of the decoder. Additionally, we obtain the downstream performance of $I^c$ by calculating the ROUGE-L score between the LLM's predicted output $\hat{y}$ (when $I^c$ is used as the instruction) against the groud-truth output $y$, denoted as $r(y,\hat{y})$. The model is  then trained with a list-wise loss to align the scores $Q_{\textrm{T5}}(I^c,x)$ of all candidate instructions with their corresponding downstream performance $r(y,\hat{y})$, while considering relative superiority among different instructions. Specifically, we first normalize both the list of predicted scores $Q_{\textrm{T5}}(I^c,x)$ and the list of downstream performance $r(y,\hat{y})$ by applying softmax across all candidate instructions, and then compute the KL-divergence between these two normalized distributions as the training loss:

% This makes the model aware of the relative quality among different instructions. In detail, we use softmax to normalize the predicted scores $Q_{\textrm{T5}}(I^c,x)$ as well as the downstream ROUGE-L scores $s(y,\hat{y})$ across different instructions. The KL-divergence between these two normalized distributions is computed as the training loss:
% In detail, we compute the KL-divergence between the normalized instruction scores $Q_{\textrm{T5}}(I^c|x)$ across the 8 sampled candidate instructions and the normalized downstream ROUGE-L scores $s(y, \hat{y})$ that depicts the performance:

%\noindent The score for the instruction is calculated as the logit of the ``yes'' token at the starting position of the decoder. The model is trained with a ranking loss to estimate the downstream performance of these instructions, while maintaining awareness of the relative quality among them. This is achieved by comparing the KL-divergence of the normalized instruction scores $Q_{\textrm{T5}}(I^c|x)$ across the 8 sampled candidate instructions against the normalized downstream ROUGE-L scores $s(y, \hat{y})$ as used to prompt the LLM:

\vspace{-0.4cm}
\begin{equation}
\nonumber
\begin{split}
    \mathcal{L} = \frac{1}{|\mathcal{B}|}\hspace{-0.05cm}\sum_{(x,y)\in\mathcal{B}}\hspace{-0.15cm}\mathbb{KL}\hspace{-0.05cm}\left(\sigma\hspace{-0.1cm}\left(r\left(y, \hat{y}\right)\right)||\hspace{+0.05cm}\sigma\hspace{-0.1cm}\left(Q_{\textrm{T5}}\left(I^c,x\right)\right)\hspace{-0.05cm}\right),\\
    \textrm{where}\ \hat{y}\sim P_{\textrm{LLM}}(\cdot|I^c, \{x^d_i, y^d_i\}^n_{i=1}, x).
    \vspace{-0.3cm}
\end{split}
\end{equation}
Note that $\mathcal{B}$ is a batch of examples and $\sigma$ is the softmax function. During testing, given a specific test example, among all candidate instructions, we select the $I^c$ that achieves the highest score $Q_{\textrm{T5}}(I^c,x)$ as the final instruction, and prompt LLM with it to obtain the desired output.
% Finally, the selected instruction is used to prompt the LLM on the same example to generate the corresponding outputs.

\subsubsection{Training Data}
\label{sec:training_data}

To train such a ranking model with generalizability to out-of-domain tasks, we categorize the tasks in the SuperNI benchmark by their task type (\textit{e.g.}, QA, sentiment analysis, etc.) and group these categories into training and test sets. We exclude tasks involving non-English languages or those with excessively long inputs. To avoid data leakage, we also exclude tasks from the training data which are sourced from the same dataset as any test task. This yields 575 tasks for training and 91 for testing. We sample up to 400 examples from each training task, which leads to 122k in total. Additional data pre-processing and filtering methods utilized to accelerate the training process can be found in Appendix~\ref{app:data_filter}.

\section{Experiments}
\label{sec:experiments}
\subsection{Settings}
To evaluate our approach under the true few-shot setting, we test it on a variety of out-of-domain tasks --- 91 from SuperNI~\cite{superNI} and 27 from BBH \cite{bbh}, where there is no overlap between task categories in training and testing. The SuperNI test set comprises both classification and generation tasks, \textit{e.g.}, commonsense classification, information extraction, etc\footnote{The full list of SuperNI test tasks is in Appendix~\ref{app:test_tasks}.}. BBH presents a diverse set of tasks spanning commonsense QA and math problems. Average ROUGE-L\footnote{The original authors of SuperNI found ROUGE-L positively correlated to accuracy on classification tasks, so average ROUGE-L is applied for simplicity.} and exact-match accuracy are used for evaluation on SuperNI and BBH, respectively. Our main experiments are conducted using OpenAI's \textit{text-davinci-003} for instruction generation and downstream inference. We also explored the instructions generated by ChatGPT (\textit{gpt-3.5-turbo}) or GPT-4~\cite{gpt-4} in \S\ref{sec:chatgpt}.

In the zero-shot setting, the ranking model is separately trained on data where downstream ROUGE scores of candidate instructions are likewise obtained under zero-shot prompting. For zero-shot classification tasks, we append additional formatting instructions to the seed instruction to narrow down the answer options in both instruction generation and downstream inference. Additional experimental settings can be found in Appendix~\ref{app:settings}.

% \usepackage{booktabs}
% \usepackage{multirow}
\begin{table*}[t]
\centering
\setlength{\tabcolsep}{1.3mm}{
\resizebox{0.98\textwidth}{!}{
\begin{tabular}{l|cc|cc|cc@{}}
\toprule
\multicolumn{1}{l|}{\multirow{2}{*}{\textbf{Methods}}} & \multicolumn{1}{c}{\multirow{2}{*}{\textbf{Generation}}} & \multicolumn{1}{c|}{\multirow{2}{*}{\textbf{Ranking}}} & \multicolumn{2}{c|}{\textbf{\textit{Few-shot}}}                          & \multicolumn{2}{c}{\textbf{\textit{Zero-shot}}}                                    \\
\multicolumn{1}{c|}{}   & \multicolumn{1}{c}{} & \multicolumn{1}{c|}{}           & \textbf{\ SuperNI\ }                   & \textbf{\ BBH\ }                        & \textbf{\ SuperNI\ }  & \textbf{\ BBH\ }                                \\ \midrule
Empty Instruction*    & None & None                  & 57.03 & 51.18 &                 35.86            & 45.12                              \\
Human Instruction*     & Human & None                & 60.94                     & 50.30                      & {46.81}   & 45.59                              \\
Random Selection\dag     & LLM & Random             & 61.61                     & 50.88                      & {45.80}   & 45.98                              \\
iPrompt*            & LLM (\textit{iterative}) & Examples                      &   57.08    &   50.46    & -                           &      -                              \\

iPrompt+*            & LLM (\textit{iterative}) & Examples                      &   61.13    &   50.82    & -                           &      -                              \\

Cross-Validation*    & LLM & Examples                & 62.02                     & 51.20                      & -                           & {-}              \\
LM Selection\dag      & LLM & LLM                  & 61.69                     & 51.96                      & {44.19}   & {45.05}          \\
On-the-fly Generation\dag    & LLM & None         & 61.03                     & 51.38                      & 45.85                       & {45.47}          \\
Auto-Instruct\dag          & LLM & Trained Model            & \textbf{64.35}                     & \textbf{52.04}                      & \textbf{49.50}              & \textbf{47.35} \\ \bottomrule
\end{tabular}}}
\caption{Results on SuperNI (91 tasks) and BBH (27 tasks) under the few-shot and zero-shot setting respectively. We report ROUGE-L on SuperNI and accuracy on BBH. Methods with * apply the same instruction for a certain task, while methods with \dag\ can select different instructions for different examples. iPrompt iteratively generates and ranks candidate instructions, while other methods adopt a generate-then-rank pipeline. We note that iPrompt, iPrompt+ and Cross-Validation are not applicable under the zero-shot setting due to the need of validation examples. Detailed results on SuperNI of different task categories can be found at Appendix~\ref{app:detail_results}.}
\label{tab:main_results}
\end{table*}

\subsection{Baselines}

As baselines in our experiments, we first consider three alternative approaches based solely on prompting the LLM:

\noindent{\textbf{(1) Cross-Validation}}. We leverage the 3-shot demonstrations as validation data to rank the instructions, with each one acting as the test example iteratively while the other two serve as demonstrations. The ROUGE-L score (or accuracy for BBH) is used as the primary ranking criterion, and the log-probability of the ground-truth output is compared as tiebreaker. The instruction selected by the demonstrations is then applied on all test examples within the same task.

\noindent{\textbf{(2) LM Selection}}. We directly prompt the LLM itself to select an instruction by enumerating all candidate instructions in a single prompt. We number the instructions and ask the LLM to generate the number of the instruction it deems most suitable to each test example.

\noindent{\textbf{(3) On-the-fly Generation}}. As a simplified variant without instruction ranking, the model is asked to directly generate an instruction for each test example. The generated instruction is then used to prompt the LLM for the same example.

Furthermore, we consider \textbf{iPrompt}~\cite{iprompt}, the existing state-of-the-art approach in optmizing instructions with LLMs. iPrompt iteratively generates instructions until it cannot find one with better performance on a validation set. To evaluate iPrompt under the true few-shot setting, we conduct its validation on the 3-shot demonstrations. Besides, since the original iPrompt generates instructions based on the examples without any task description, for a fair comparison, we implement an \textbf{iPrompt+} baseline that uses a similar meta-prompt to ours with the seed instruction (See Appendix~\ref{app:iprompt} for details). In addition, we evaluate the performance of not using any instruction (\textbf{Empty Instruction}), directly using the human-written seed instruction (\textbf{Human Instruction}) or randomly selecting an instruction from the generated candidates (\textbf{Random Selection}) on each task.

% \subsection{Experimental Settings}

% As outlined in \S\ref{sec:generation}, we utilize 7 distinct meta-prompts (3 groups of demonstration tasks + 4 heuristic style-specific prompts) for instruction generation. We remove the meta-prompt about explaining demonstrations in the zero-shot setting. We sample 3 candidate instructions for each meta-prompt, resulting in 22 candidate instructions for the few-shot setting and 19 for zero-shot experiments after including the seed instruction into ranking.

% When training the instruction ranking model, we sample up to 400 examples from each training task, which leads to 122k in total.

\subsection{Results}

The overall results of SuperNI and BBH are shown in Table~\ref{tab:main_results}, where scores are averaged across all tasks. Auto-Instruct shows notable consistency and generalizability in out-of-domain scenarios, surpassing all baselines across different benchmarks and settings. Key findings are outlined below.

\vspace{0.2cm}
\noindent{\textbf{The LLM shows competitive ability in generating effective instructions, yet ranking is still necessary.}} In alignment with previous work~\cite{human_level_instruction_engineer, iprompt}, the LLM is able to generate effective instructions for various tasks. Our style-specific meta-prompts enables the LLM to produce a diverse set of instructions to cater to varied scenarios where different tasks may favor different styles of instructions. In the few-shot setting, the LLM-generated instructions already surpass their human-written counterparts on average, as indicated by the random selection scores. Although humans may have prior knowledge of some examples when they write the instructions, the LLM, not given any demonstrations in the zero-shot setting, generates instructions of comparable quality to those written by humans. Nevetheless, neither random selection nor directly generating a single instruction (\textit{i.e.}, on-the-fly generation) significantly improves over the human-written baseline. This aligns with the instability of the LLM performance across different instructions as discussed in Figure~\ref{fig:instruction_stdev}, which indicates further instruction ranking is still essential.

\vspace{0.1cm}
\noindent{\textbf{Simply prompting the LLM or using the validation data are not reliable in the low-resource setting.}} Although offering the convenience of not training any models, both directly prompting the LLM (LM selection) and using few-shot demonstrations for validation (iPrompt and cross-validation) fail to deliver consistently improved results compared to random selection. This highlights that (1) the LLM itself lacks clue of the expected downstream performance of different instructions; (2) the volume of validation data must be substantial enough to effectively estimate the performance of instructions on the test data, which brings high cost in many realistic scenarios.

\vspace{0.1cm}
\noindent{\textbf{Our trained instruction ranking model is the most effective approach to select instructions so far.}} Although the data and instructions for out-of-domain tasks are not seen by the ranking model, it exhibits promising generalizability in selecting effective instructions thanks to the training on hundreds of different tasks. For example, on the SuperNI benchmark, it outperforms random selection by 4\% and 8\% on few-shot and zero-shot settings respectively. Besides, our complete pipeline delivers a relative 6\% improvement over the original human instructions in both few-shot and zero-shot settings, indicating that the human-written instructions still need improvement in many contexts.

% \usepackage{booktabs}
\begin{table}[t]
\centering
\resizebox{0.44\textwidth}{!}{
\begin{tabular}{l|cc}
\toprule
\multicolumn{1}{l|}{\textbf{Methods}}         & 
\hspace{0.5cm} \textbf{ChatGPT}          & \textbf{GPT-4}           \\ \midrule
\multicolumn{3}{c}{\textit{Few-shot, instructions from text-davinci-003}} \\ \midrule
\multicolumn{1}{l|}{Human}           & \hspace{0.5cm}60.39            & 67.31           \\
\multicolumn{1}{l|}{Random}          & \hspace{0.5cm}60.44            & 67.07           \\
\multicolumn{1}{l|}{Auto-Instruct}   & \hspace{0.5cm}\textbf{62.88}   & \textbf{69.45}  \\ \midrule
\multicolumn{3}{c}{\textit{Few-shot, instructions from} ChatGPT/GPT-4}             \\ \midrule
\multicolumn{1}{l|}{Human}           & \hspace{0.5cm}60.39            & 67.31           \\
\multicolumn{1}{l|}{Random}          & \hspace{0.5cm}60.44            & 66.77           \\
\multicolumn{1}{l|}{Auto-Instruct}   & \hspace{0.5cm}\textbf{62.32}   & \textbf{68.16}  \\ \midrule
\multicolumn{3}{c}{\textit{Zero-shot, instructions from} ChatGPT/GPT-4}            \\ \midrule
\multicolumn{1}{l|}{Human}           & \hspace{0.5cm}47.77            & 54.11           \\
\multicolumn{1}{l|}{Random}          & \hspace{0.5cm}46.22            & 53.06           \\
\multicolumn{1}{l|}{Auto-Instruct}   & \hspace{0.5cm}\textbf{49.04}   & \textbf{55.53}  \\ \bottomrule
\end{tabular}}
\caption{SuperNI results of transferring Auto-Instruct to ChatGPT and GPT-4, using either (1) instructions generated by \textit{text-davinci-003}, or (2) instructions generated by the same model as downstream inference (\textit{i.e.}, ChatGPT or GPT-4). The instruction ranking model is still the one trained on \textit{text-davinci-003} instructions.}
% \caption{Results after replacing \textit{text-davinci-003} with ChatGPT or GPT-4 in both instruction generation and downstream inference on SuperNI. Our model is still able to rank ChatGPT or GPT-4 instructions.}
\label{tab:chatgpt}
\end{table}
\begin{table}[t]
\setlength{\tabcolsep}{1.3mm}{
\resizebox{0.49\textwidth}{!}{
\begin{tabular}{@{}l|cc|cc@{}}
\toprule
\multirow{2}{*}{\textbf{Methods}} & \multicolumn{2}{c|}{\textbf{Selection Acc}} & \multicolumn{2}{c}{\textbf{Win Rate}} \\ 
                         & Top1              & Top5             & \textit{vs.}\hspace{0.1cm}Empty        & \textit{vs.}\hspace{0.1cm}Human        \\ \midrule
Human       &   45.25   &   70.35   &   22.43   &   - \\
Random                   & 46.76                & 70.13               & 24.95             & 16.87             \\
Cross-Validation         & 47.61                & 68.73               & 26.77             & 20.74             \\
LM Selection             & 47.53                & 71.07               & 25.17             & 17.93             \\
Auto-Instruct            & \textbf{52.54}       & \textbf{73.10}      & \textbf{29.51}    & \textbf{23.89}    \\ \bottomrule
\end{tabular}}}
\caption{Evaluation of instruction ranking on silver labels. \textit{Left}: we evaluate the percentage of cases where the selected instruction is the best (top-1) or is among top-5 candidates, according to the actual downstream performance. We note that there could be multiple instructions sharing the best score. \textit{Right}: we check the percentage of selected instructions that outperform either the empty instruction or the human-written ones.}
\label{tab:ranking}
\end{table}

\subsection{Analysis}

In this section, we delve deeper into the performance of our approach by analyzing the use of other LLMs for instruction generation, the performance on seen tasks, the size of training data, and case studies. Additional analysis of the comparison between Auto-Instruct and multi-answer ensemble is in Appendix~\ref{app:experiments}. These analyses are conducted in the few-shot setting unless stated otherwise.

\subsubsection{Generalization to other LLMs}
\label{sec:chatgpt}

To further test the generalizability of our approach, we transfer Auto-Instruct to other LLMs by using ChatGPT (\textit{gpt-3.5-turbo}) and GPT-4 as downstream inference models. As Table~\ref{tab:chatgpt} suggests, instructions selected by Auto-Instruct on \textit{text-davinci-003} are still effective if transferred to ChatGPT and GPT-4. Furthermore, our instruction ranking model is able to rank instructions generated by ChatGPT or GPT-4 under both few-shot and zero-shot scenarios, despite not having seen any instruction created by these LLMs during training. Improved results can also be seen when transferring Auto-Instruct to LLaMA-2-chat~\cite{llama-2}, a recent open-source LLM, as shown in Appendix~\ref{app:llama-2}. As a conclusion, despite variations in phrasing across instructions generated by different LLMs, the underlying pattern determining instruction effectiveness is transferable, although the largest improvement is still seen in the same-LLM experiments. Suffice to say, our trained instruction ranking model can be directly applied to select instructions for other LLMs without the need of re-training.

% To further test the generalizability of our approach, we generate instructions for SuperNI using ChatGPT (\textit{gpt-3.5-turbo}) and GPT-4. To be specific, we substitute the LLM used for both instruction generation and downstream inference with either ChatGPT or GPT-4. Interestingly, although the ranking model is only trained on instructions scored by \textit{text-davinci-003}, it is able to rank instructions generated by ChatGPT or gpt-4, as shown in Table~\ref{tab:chatgpt}. This suggests that despite variations in phrasing across instructions generated by different LLMs, the underlying pattern of instruction effectiveness is transferable, although the largest improvement is still seen in the same-LLM experiments.

\subsubsection{Evaluation of Instruction Ranking}

To investigate the effectiveness of the instruction ranking model, we compare it with other instruction selection baselines by assigning silver labels to candidate instructions, with results detailed in Table~\ref{tab:ranking}. First, we use the actual downstream performance of the candidate instructions as silver labels. Our ranking model is more capable of distinguishing better instructions, as shown by an evidently higher accuracy of picking the top-1 or top-5 instructions among all 22 candidates. Second, we evaluate how often the selected instruction improves the downstream performance in comparison to either the empty instruction or the human-written instruction. Once again, the instructions from our ranking model makes the most significant improvements, advancing the human-written counterparts in 7\% more cases than random selection. The consistent performance gain across all silver-label evaluations further corroborates the superiority of our model over alternative ranking methods based on cross-validation or LM selection.

% \usepackage{booktabs}
\begin{table}[t]
\centering
\resizebox{0.44\textwidth}{!}{
\begin{tabular}{@{}l|cc@{}}
\toprule
\textbf{Methods}           & \textbf{Unseen Tasks} & \textbf{Seen Tasks} \\ \midrule
Human                     & 54.59                 & 40.32               \\
Random                    & 55.57                 & 39.74               \\
Auto-Instruct                 & \textbf{60.18}        & \textbf{45.89}      \\
~$\vdash$ (\textit{vs.} Random) & \textbf{(+8.3\%)}        & \textbf{(+15.5\%)}     \\ \bottomrule
\end{tabular}}
\caption{Results on instruction-sensitive test data of both seen tasks (100 tasks seen in training) and unseen tasks (the same as Table~\ref{tab:main_results}) from SuperNI. We additionally report the relative improvement ratio to the random selection baseline since the vanilla performance is not on the same scale.}
\vspace{-0.3cm}
\label{tab:seen_tasks}
\end{table}

\begin{figure}[t]
    \centering
    \includegraphics[width=0.4\textwidth]{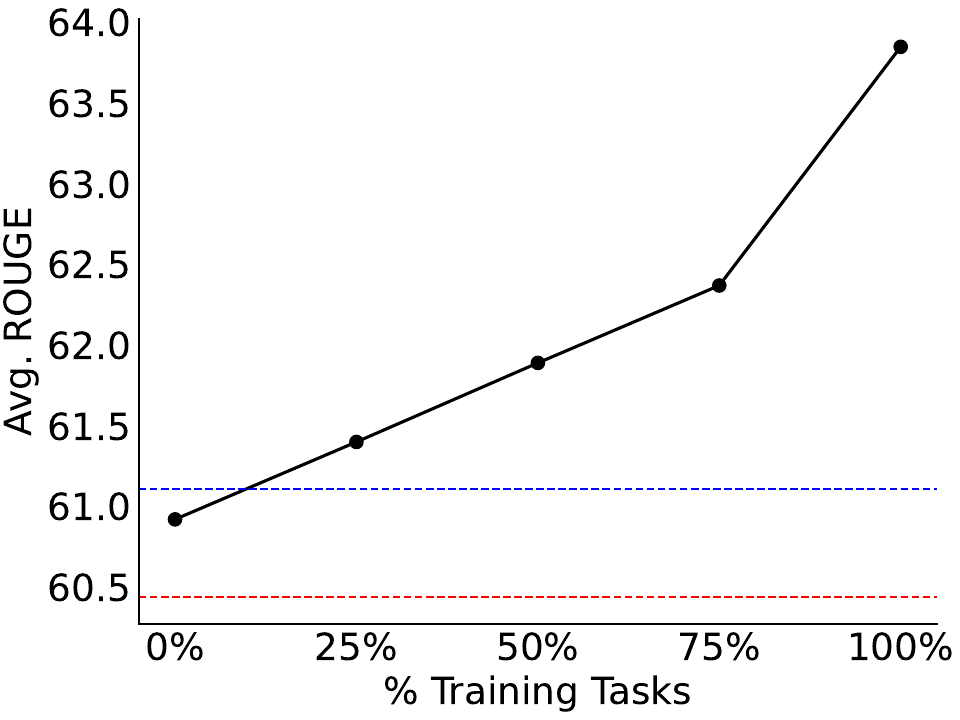}
    \caption{Results of using different number of training tasks. 0\% means directly using the pre-trained FLAN-T5 checkpoint in instruction ranking, which shows a similar performance to random instruction selection.}
    \label{fig:training_tasks}
\end{figure}

\subsubsection{Auto-Instruct on Seen Tasks}

Besides the out-of-domain setting, we explore an in-domain setting where we select additional examples from tasks seen during training, so as to further examine the competency of the instruction ranking model. For a fair comparison of the model's ranking abilities across different tasks, we experiment with instruction-sensitive examples, defined as examples where not all candidate instructions yield the same ROUGE score. We sample 100 additional examples from each of 100 tasks that are seen in training but not included in the dev set. As presented in Table~\ref{tab:seen_tasks}, the model shows enhanced ranking ability on seen tasks due to prior exposure to the instructions during training. This indicates that our approach is useful in both data-rich and data-scarce circumstances.

\begin{figure*}[t]
    \centering
    \includegraphics[width=1.0\textwidth]{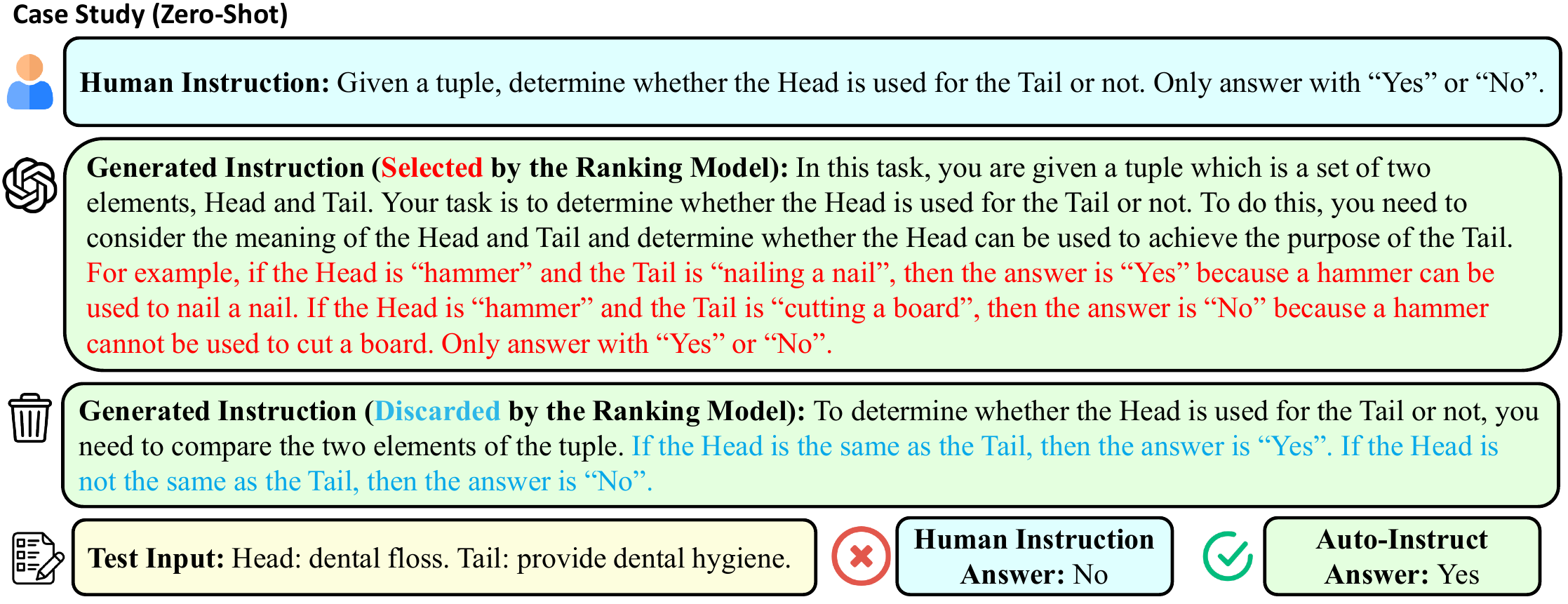}
    \vspace{-0.5cm}
    \caption{In this case, Auto-Instruct selects an instruction which ``transforms'' the zero-shot inference to a ``2-shot'' inference by providing additional examples (highlight in {\color{red}red}), while discarding an instruction that includes hallucination in the task description (highlight in \textcolor[RGB]{45, 183, 232}{blue}). The human instruction is also included in ranking candidates.}
    \vspace{-0.2cm}
    \label{fig:case_study}
\end{figure*}

\begin{figure}[t]
    \centering
    \includegraphics[width=0.48\textwidth]{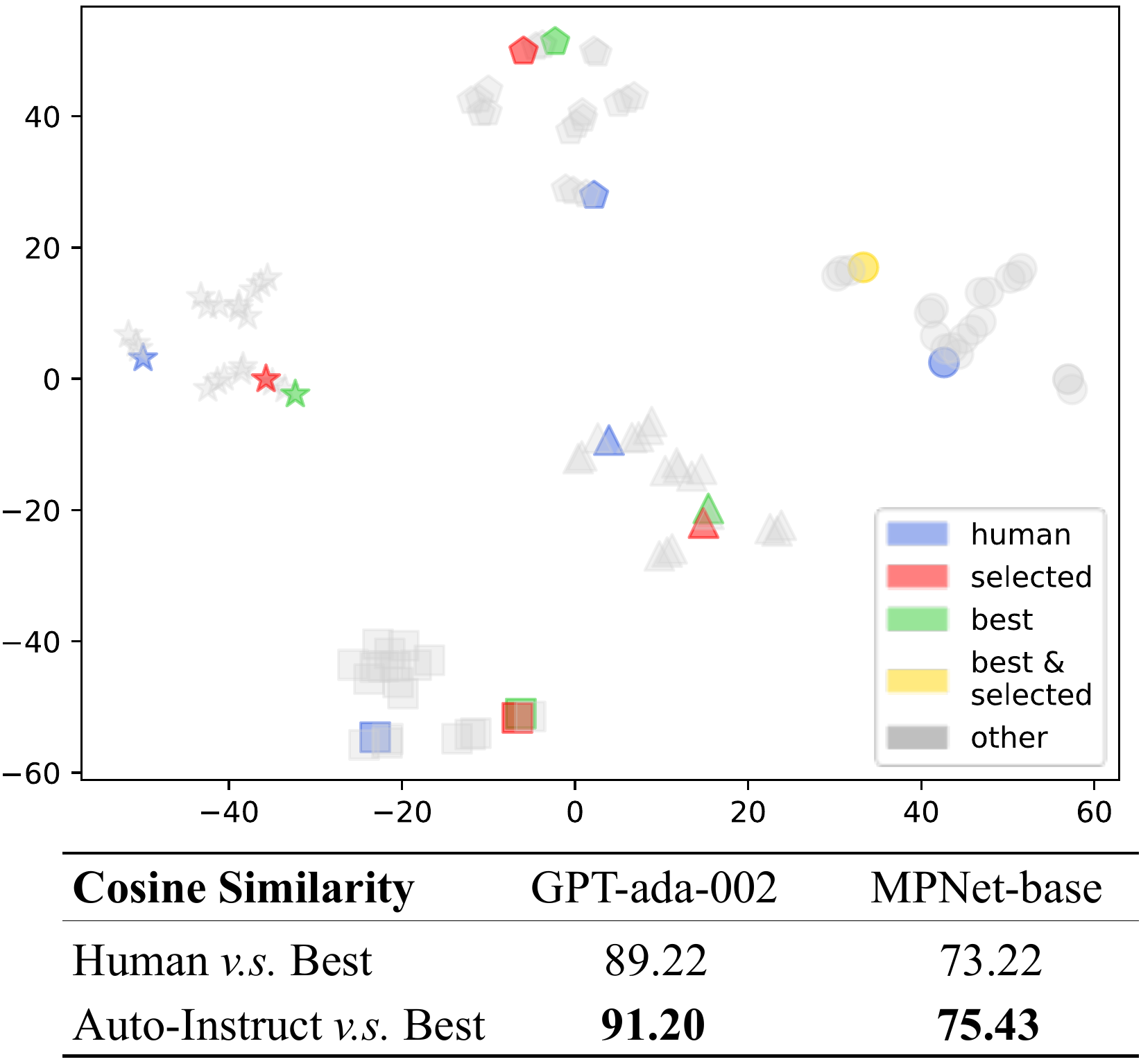}
    \vspace{-0.5cm}
    \caption{Above: Instruction embeddings of 5 SuperNI tasks where Auto-Instruct selected instruction performs better than human instruction, as visualized by T-SNE. ``Best'' refers to the instruction with the highest ROUGE score. Below: Average cosine similarity between instruction embeddings on all SuperNI tasks. Two embedding models are \textit{text-embedding-ada-002} from OpenAI and \textit{all-mpnet-base-v2} from Sentence-Transformers\footnotemark. Best viewed in color.}
    \label{fig:embeddings}
    \vspace{-0.2cm}
\end{figure}

\subsubsection{Effect of More Training Tasks}

To analyze the effect of large-scale multi-task training on out-of-domain generalizability, we manipulate the number of training tasks of the instruction ranking model. Specifically, We exclude tasks from the training set by their category, \textit{i.e.}, all tasks from selected categories are removed. As shown in Figure~\ref{fig:training_tasks}, the increment in the number of training tasks from additional categories is a key contributor to the superior performance of our model compared to the random selection baseline. Since the performance has not plateaued when all tasks are included, it is plausible to expect further performance gains if more training tasks are available.

\subsubsection{Analysis of the Selected Instructions}
\label{sec:case_study}

Figure~\ref{fig:embeddings} illustrates how our selected instructions improve the original human instructions. As indicated by the average similarity scores, Auto-Instruct is able to provide instructions more similar to the optimal ones among the candidates.\footnotetext{\url{www.sbert.net/docs/pretrained_models.html}} As demonstrated by the scatter plot, in scenarios where the selected instruction outperforms the human instruction, its embedding usually deviates significantly from that of the human instruction but stays close to the optimal one. These results suggest that the selected instruction refines the human-written seed instruction by progressing towards the ideal solution, while the embedding distance between the selected and seed instructions makes such improvement hard to achieve by pure manual engineering.

In addition, we offer a case study in Figure~\ref{fig:case_study} in the zero-shot setting where the LLM cannot refer to any demonstrations. Nevertheless, the LLM manages to generate additional examples using the knowledge gained from its extensive pre-training. These additional examples can act as demonstrations to create a ``2-shot inference'' setting, leading to a correct prediction that could not be achieved via the original zero-shot inference. Conversely, we also present an example where the LLM-generated instruction includes hallucinated descriptions that distort the original meaning of the seed instruction. The mismatch between this instruction and the test example results in its rejection by the ranking model. Readers may find further case studies in Appendix~\ref{app:case_study}.

% In Figure~\ref{fig:case_study}, We present some instructions generated and selected by our Auto-Instruct pipeline which provides positive impact to solving the task. In the first case from the few-shot setting, the human instruction provides a general and concise description of what the input question is asking for. On the other hand, Auto-Instruct offers a more concrete instruction on what information is provided in the input and which part should be paid attention to. Moreover, it also provides the answer reasoning process of the demonstration examples which can shed light on how to solve the test example. In the second case from the zero-shot setting, compared to the original human instruction, not only does Auto-Instruct provide a more detailed explanation, but it also generates some examples which can serve similarly as demonstrations. This provides valuable information for the LLM to answer the question when no demonstrations are available.

\section{Conclusion}
In this work, we introduce Auto-Instruct, an automatic approach of generating, ranking and selecting instructions, which offers a solution to the high cost and subjectivity associated with human-engineered instructions. Our approach begins by prompting the LLM to generate a diverse set of candidate instructions. Next, an instruction ranking model trained on hundreds of tasks is used to rank the candidate instructions and select the most effective one to solve a specific example. Experimental results demonstrate that our approach provides better instructions than both human-written ones and those produced by previous instruction generation approaches, as tested on 118 out-of-domain tasks.

\section*{Limitations}
To our knowledge, this work has the following limitations:
\begin{itemize}
    \item Due to the considerable cost associated with OpenAI models, and the limited capacity of their API interface, we only score the candidate instructions on a moderate number of tasks as described in \S\ref{sec:training_data}. Given the results in Figure~\ref{fig:training_tasks}, we expect that the model could demonstrate improved generalizability if more training data with labeled instructions were available.
    \item The scope of this study is limited to the generation of instructions in English; tasks in non-English languages are not part of our training data. As a result, the model might not perform satisfactorily for non-English tasks. Further investigation into generating cross-lingual instructions is left for future work.
    \item Despite employing a wide range of meta-prompts, which significantly mitigates the dependence on prompt engineering, the phrasing of these meta-prompts could still influence the quality of the instructions generated. We leave the exploration of automatically diversify the generated instructions as future work.
\end{itemize}

\nocite{liang2023mathtutor}
\nocite{zhihan_ODQA}
\nocite{edmem}
\nocite{mengxia}

% EMNLP 2023 requires all submissions to have a section titled ``Limitations'', for discussing the limitations of the paper as a complement to the discussion of strengths in the main text. This section should occur after the conclusion, but before the references. It will not count towards the page limit.  

% The discussion of limitations is mandatory. Papers without a limitation section will be desk-rejected without review.
% ARR-reviewed papers that did not include ``Limitations'' section in their prior submission, should submit a PDF with such a section together with their EMNLP 2023 submission.

% While we are open to different types of limitations, just mentioning that a set of results have been shown for English only probably does not reflect what we expect. 
% Mentioning that the method works mostly for languages with limited morphology, like English, is a much better alternative.
% In addition, limitations such as low scalability to long text, the requirement of large GPU resources, or other things that inspire crucial further investigation are welcome.

% \section*{Ethics Statement}
% Scientific work published at EMNLP 2023 must comply with the \href{https://www.aclweb.org/portal/content/acl-code-ethics}{ACL Ethics Policy}. We encourage all authors to include an explicit ethics statement on the broader impact of the work, or other ethical considerations after the conclusion but before the references. The ethics statement will not count toward the page limit (8 pages for long, 4 pages for short papers).

\section*{Acknowledgements}
This work was supported by NSF IIS-2119531, IIS-2137396, IIS-2142827, IIS-2234058, CCF-1901059, and ONR N00014-22-1-2507. We thank Canwen Xu (University of California San Diego) for his valuable suggestions during paper writing.
% This document has been adapted by Yue Zhang, Ryan Cotterell and Lea Frermann from the style files used for earlier ACL and NAACL proceedings, including those for 
% ACL 2020 by Steven Bethard, Ryan Cotterell and Rui Yan,
% ACL 2019 by Douwe Kiela and Ivan Vuli\'{c},
% NAACL 2019 by Stephanie Lukin and Alla Roskovskaya, 
% ACL 2018 by Shay Cohen, Kevin Gimpel, and Wei Lu, 
% NAACL 2018 by Margaret Mitchell and Stephanie Lukin,
% Bib\TeX{} suggestions for (NA)ACL 2017/2018 from Jason Eisner,
% ACL 2017 by Dan Gildea and Min-Yen Kan, NAACL 2017 by Margaret Mitchell, 
% ACL 2012 by Maggie Li and Michael White, 
% ACL 2010 by Jing-Shin Chang and Philipp Koehn, 
% ACL 2008 by Johanna D. Moore, Simone Teufel, James Allan, and Sadaoki Furui, 
% ACL 2005 by Hwee Tou Ng and Kemal Oflazer, 
% ACL 2002 by Eugene Charniak and Dekang Lin, 
% and earlier ACL and EACL formats written by several people, including
% John Chen, Henry S. Thompson and Donald Walker.
% Additional elements were taken from the formatting instructions of the \emph{International Joint Conference on Artificial Intelligence} and the \emph{Conference on Computer Vision and Pattern Recognition}.

\bibliography{reference}
\bibliographystyle{acl_natbib}

\newpage
\appendix

\section{Training Data Pre-Processing}
\label{app:data_filter}

As detailed in \S\ref{sec:ranking}, the instruction ranking model is trained to rank candidate instructions against their downstream performance on the LLM. The downstream performance of an instruction $I^c$ refers to how well the LLM's predicted output $\hat{y}$ matches the ground-truth output $y$ when using $I^c$ to prompt the LLM, as quantified by the ROUGE-L score $r(y,\hat{y})$. To calculate this score, we pair each training example with all 22 candidate instructions of the corresponding task (generated with the method in \S\ref{sec:generation}), and collect the LLM's predicted output to the example prompted by each candidate instruction. After calculating the ROUGE-L scores against the ground-truth, we discard examples where candidate instructions are not distinctly rankable -- in cases where the range of downstream performance across different instructions is less than 10 points in ROUGE-L.

To accelerate the training process, we sample 8 candidate instructions from the total pool of 22 for each example, and train the model to rank these 8 instructions. However, in some tasks, certain instructions may significantly outperform others. Uniformly sampling 8 candidate instructions could result in such ``extraordinary'' instructions being disproportionately favored too many times in the training of the ranking model. To address this, we inversely proportion the sampling rate of each instruction to its popularity (\textit{i.e.}, the number of cases where this instruction is superior to all others). Finally, we sample up to 400 examples from each training task, which leads to 122k training examples in total.

% In practice, to expedite the training process, we sample 8 candidate instructions out of the total 22 instructions for each example and train the model to rank these 8 instructions. Additional data pre-processing strategies can be found in Appendix~\ref{app:data_filter}. Finally, we sample up to 400 examples from each training task, which leads to a total of 122k examples and each is paired with 8 candidate instructions.

% To obtain the training data on the effectiveness of different instructions, we score all candidate instructions according to their performance on each training example. Each candidate instruction is scored for a given example using the ROUGE-L score between the ground-truth output and the LLM prediction when prompted by this instruction. To expedite training, we sample 8 instructions from all candidate instructions for each example. To avoid any ``extraordinary'' instruction dominating the training data on a certain task, we inversely proportion the sampling rate of each instruction to its popularity (\textit{i.e.}, the number of examples where this instruction yields the best results among all instructions). Finally, we discard examples where the variation across all instructions is less than 10 points in ROUGE-L to ensure a strong and reliable training signal.

\section{Detailed Experimental Settings}
\label{app:settings}

The instruction ranking model is initialized with FLAN-T5-Large (780M parameters; \citealp{flan-t5}), and is trained using Adafactor~\cite{adafactor} with learning rate 5e-5, batch size 128 and dropout rate 0.1. We employ an in-domain dev set including a total of 5k unseen examples from 100 training tasks to select the best checkpoint within 5 epochs. The validation performance on the dev set is 67.66 in ROUGE-L, while random selection only achieves a score of 54.28. When using OpenAI models, for instruction generation, we set the maximum instruction length to be 300 tokens, and we use a temperature of 1.0 and \texttt{top\_p} of 0.75 for token sampling; for downstream inference, we set both to 0 for deterministic outputs. Generating all candidate instructions for 91 SuperNI test tasks cost us 18 USD in total, according to OpenAI's pricing (0.02 USD per 1k tokens for \textit{text-davinci-003}). In \textit{text-davinci-003} experiments, the random selection score is calculated as the average score across all instructions on each example, including the human-written seed instruction. In ChatGPT and GPT-4 instructions, due to the limited capacity of their API interfaces, we randomly sample an instruction for each example and test its performance.

\begin{figure}[t]
    \centering
    \includegraphics[width=0.35\textwidth]{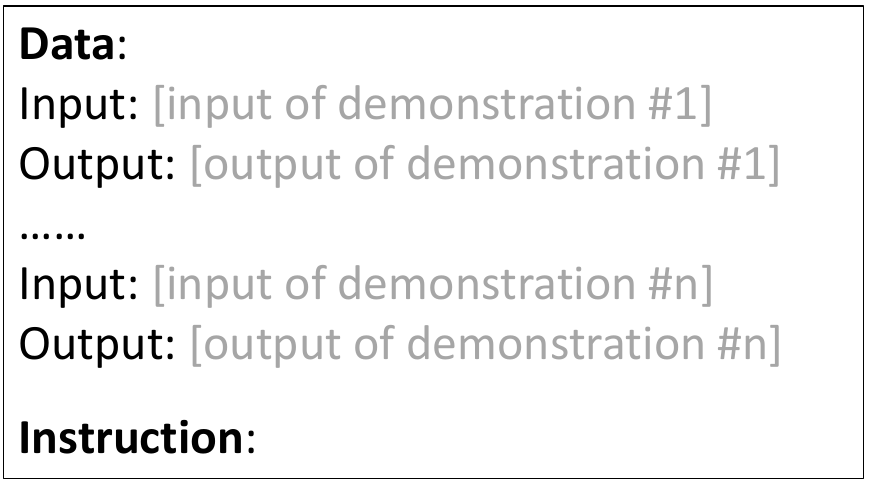}
    \caption{The meta-prompt of instruction generation with iPrompt\footnotemark.}
    \label{app_fig:iprompt}
\end{figure}

\begin{figure}[t]
    \centering
    \includegraphics[width=0.35\textwidth]{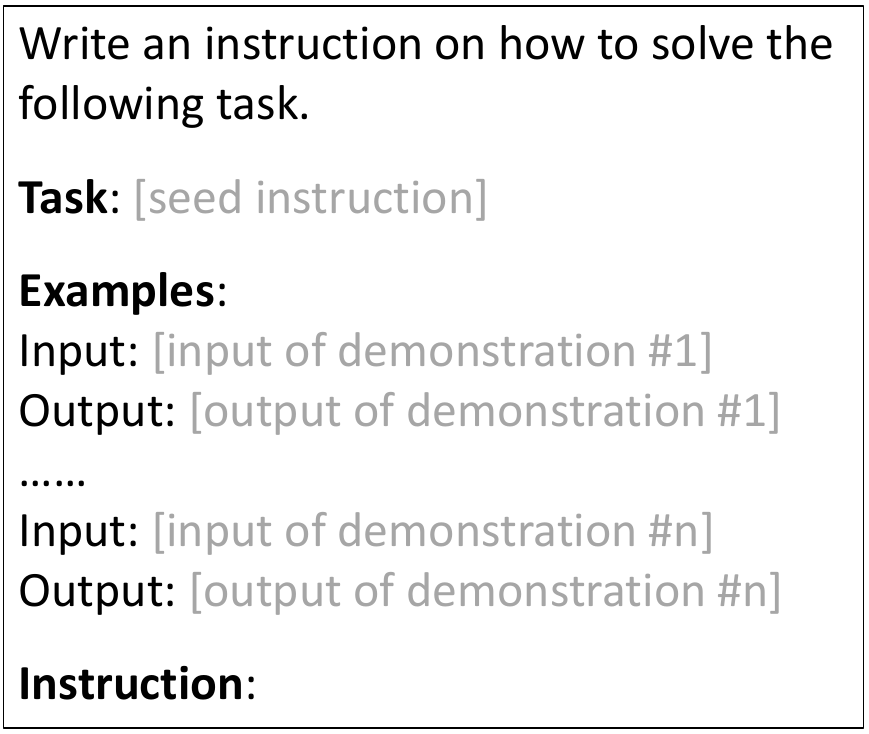}
    \caption{The meta-prompt of instruction generation with iPrompt+, similar to ours in Figure~\ref{app_fig:all-meta-prompts}.}
    \label{app_fig:iprompt_plus}
\end{figure}

\begin{figure*}
    \centering
    \includegraphics[width=0.85\textwidth]{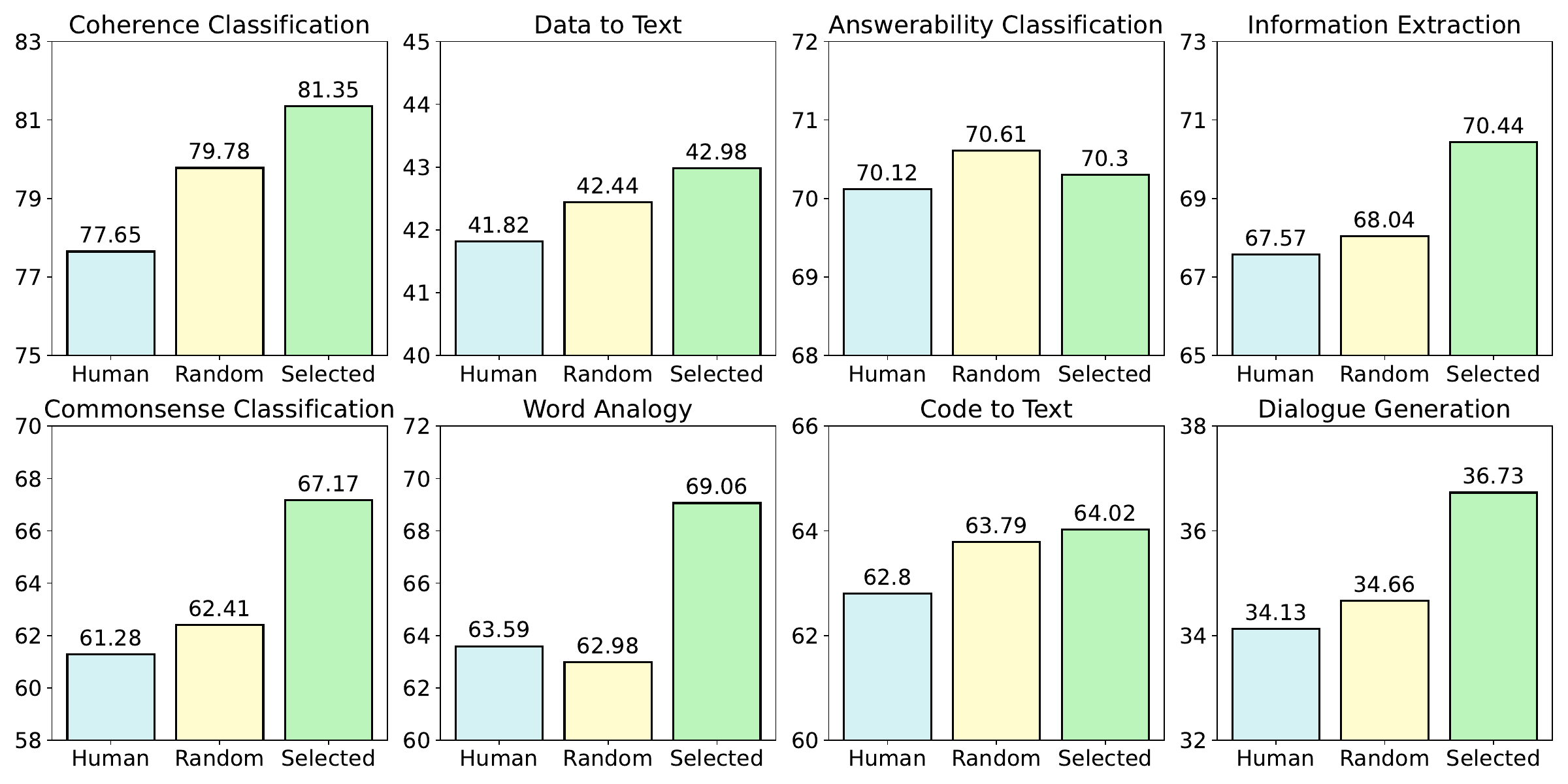}
    \caption{Few-shot performance of instructions selected by Auto-Instruct (denoted as ``Selected'') on all 8 categories of SuperNI test tasks, compared to human-written and random selected instructions.}
    \label{app_fig:task_category}
\end{figure*}

\footnotetext{In the original iPrompt implementation, the meta-prompt ends with the suffix \texttt{Prompt:}. However, this leads to incoherent instruction generation on our benchmarks. Therefore, we changed it to \texttt{Instruction:} which addressed this issue.}

\section{The iPrompt Baseline}
\label{app:iprompt}

In this section, we outline the adaptations made to the iPrompt\footnote{\url{www.github.com/csinva/imodelsX/tree/master/imodelsx/iprompt}}~\cite{iprompt} method for our setting. We mainly address two discrepancies between its original implementation and our setup: (1) the original iPrompt generates instructions using GPT-J~\cite{gpt-j}, and (2) it uses a validation set to score and select instructions. To address (1), we use \textit{text-davinci-003} for its instruction generation,  identical to the model used for downstream inference. For (2), we conduct its instruction validation on the 3-shot demonstrations. Due to the cost of iteratively requesting the OpenAI API, we incorporate an early stopping criterion which halts the process if the validation performance\footnote{The average ROUGE-L score between the LLM's predicted output and the ground-truth on validation data.} has not improved for 10 iterations. Actually, almost all tasks stopped before 30 iterations. Following this, We select the instruction with the best validation performance to evaluate on the test examples. 

According to the original codebase, we use the meta-prompt shown in Figure~\ref{app_fig:iprompt} for instruction generation with iPrompt. Since this meta-prompt does not utilize any task description, for a fair comparison, we implement an iPrompt+ baseline with a similar meta-prompt to our method which utilizes the seed instruction, as shown in Figure~\ref{app_fig:iprompt_plus}. Readers can refer to the original paper~\cite{iprompt} for technical details of iPrompt.

\section{Additional Experimental Results}
\label{app:experiments}

In this section, we present more experimental results in addition to those analyzed in Section~\ref{sec:experiments}. All experiments in this section are conducted in the few-shot setting unless stated otherwise.

\subsection{SuperNI Results by Task Category}
\label{app:detail_results}
Here, we present the detailed experimental results on 8 different categories of SuperNI test tasks (see Appedix~\ref{app:test_tasks} for the list of test tasks). As shown in Figure~\ref{app_fig:task_category}, Auto-Instruct surpasses the human-written and random instructions no matter it is evaluated on classification, extraction or generation tasks, with the only exception as answerability classification. Notably, Auto-Instruct outperforms the original human-written instruction by 10\%, 9\% and 8\% on commonsense classification (classification tasks), word analogy (short generation tasks) and dialogue generation (long generation tasks), respectively.

\subsection{Generalization to Other LLMs}
\label{app:llama-2}
% \usepackage{booktabs}
\begin{table}[t]
\centering
\resizebox{0.4\textwidth}{!}{
\begin{tabular}{l|c}
\toprule
\multicolumn{1}{l|}{\textbf{Methods}}         & 
 \textbf{LLaMA-2-chat-7B}                     \\ \midrule
\multicolumn{2}{c}{\textit{Few-shot, instructions from text-davinci-003}} \\ \midrule
\multicolumn{1}{l|}{Human}           &  53.87         \\
\multicolumn{1}{l|}{Random}          & 54.18      \\
\multicolumn{1}{l|}{Auto-Instruct}   & \textbf{55.90}  \\
\bottomrule
\end{tabular}}
\vspace{-0.1in}
\caption{SuperNI results of transferring Auto-Instruct to LLaMA-2-chat-7B, using instructions generated by \textit{text-davinci-003}. The instruction ranking model is still the one trained on \textit{text-davinci-003} instructions.}
\label{app_tab:llama-2}
\end{table}

In addition to Section~\ref{sec:chatgpt}, we further assess the generalizability of Auto-Instruct to open-source LLMs. As demonstrated in Table~\ref{app_tab:llama-2}, instructions selected by Auto-Instruct enhance the performance of LLaMA-2-chat~\cite{llama-2}. This once again underscores the capability of Auto-Instruct to generalize across different LLMs without re-training the instruction ranking model. It is worth noting that we use instructions generated by \textit{text-davinci-003} in these experiments, because both the 7B and 13B versions of LLaMA-2-chat exhibit weaker abilities in following our meta-prompts for instruction generation, contrasted with mega-size GPT models. We leave the study of instruction generation with recent open-source LLMs as future work.

\subsection{Compare to Answer Ensemble}
\label{app:self-consist}

Given that Auto-Instruct includes sampling multiple candidate instructions before selecting the best one, we compare it to another sampling approach, \textit{i.e.}, sampling and ensembling multiple answers. Using the original human-written instruction, we sample responses 10 times with nucleus sampling~\cite{nucleus}, without sampling multiple instructions. Then, we ensemble all 10 responses by marginalizing the LM probability of each unique response before selecting the most probable one, similar to the idea of self-consistency~\cite{self-consistency}. The results, shown in Table~\ref{tab:self-consist}, indicate that the answer ensemble approach only brings a marginal improvement on SuperNI, which is not comparable to the performance gain achieved with Auto-Instruct.

\begin{table}[t]
\centering
\resizebox{0.3\textwidth}{!}{
\begin{tabular}{l|c}

\toprule
\textbf{Method}  &  \textbf{Score} \\ \midrule
Human            &  60.94          \\
Human (Ensemble) &  61.08          \\
Auto-Instruct    & \textbf{64.35} \\ \bottomrule
\end{tabular}}
\caption{Results of multi-answer ensemble prompted by human-written instructions on SuperNI test tasks.}
\label{tab:self-consist}
\end{table}

\section{Meta-Prompts for Instruction Generation}
\label{app:meta_prompt}

\begin{figure*}[t]
  \centering
  \begin{subfigure}[b]{0.32\textwidth}
    \centering
    \includegraphics[width=\textwidth]{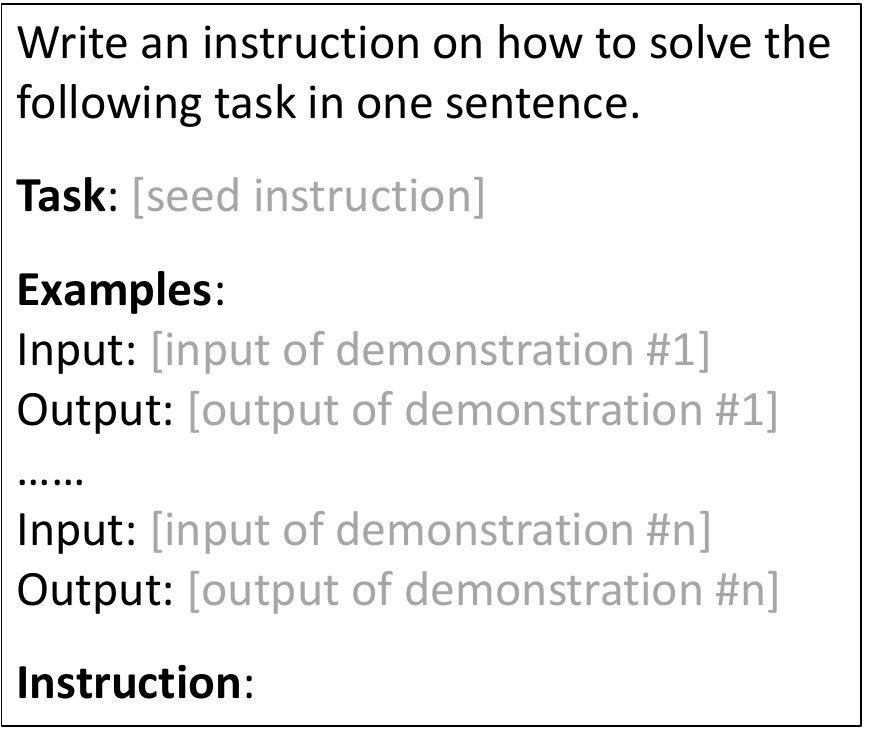}
    \caption{One-sentence instruction}
    \label{fig:one-sentence}
  \end{subfigure}
  \hfill
  \begin{subfigure}[b]{0.32\textwidth}
    \centering
    \includegraphics[width=\textwidth]{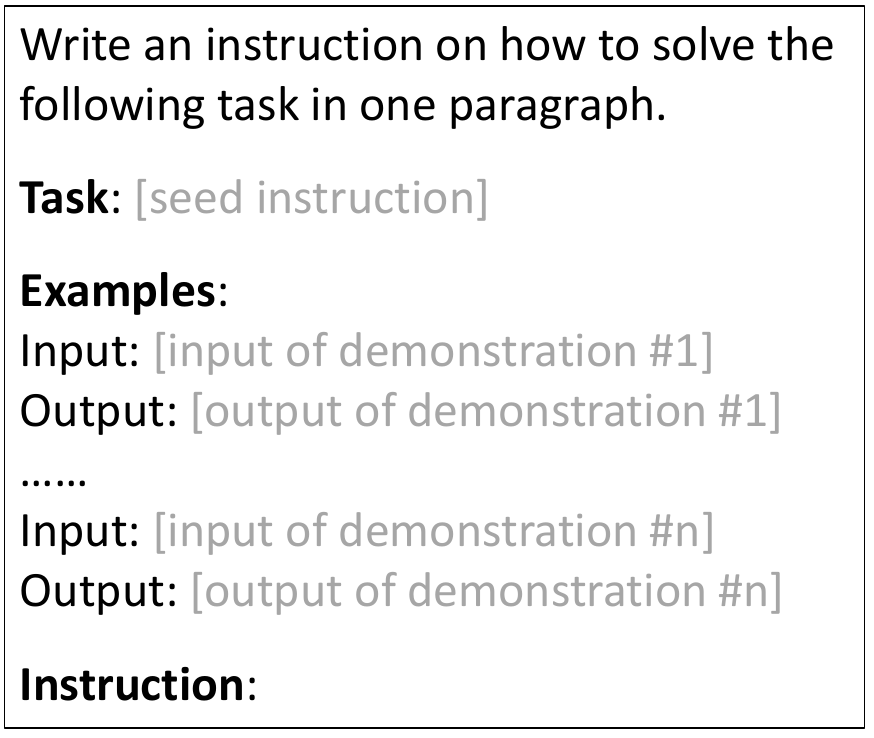}
    \caption{One-paragraph instruction}
    \label{fig:one-paragraph}
  \end{subfigure}
  \hfill
  \begin{subfigure}[b]{0.32\textwidth}
    \centering
    \includegraphics[width=\textwidth]{Figures/step_by_step.pdf}
    \caption{Step-by-step instruction}
    \label{fig:step-by-step}
  \end{subfigure}
  \vskip\baselineskip
  \begin{subfigure}[b]{0.48\textwidth}
    \centering
    \includegraphics[width=0.7\textwidth]{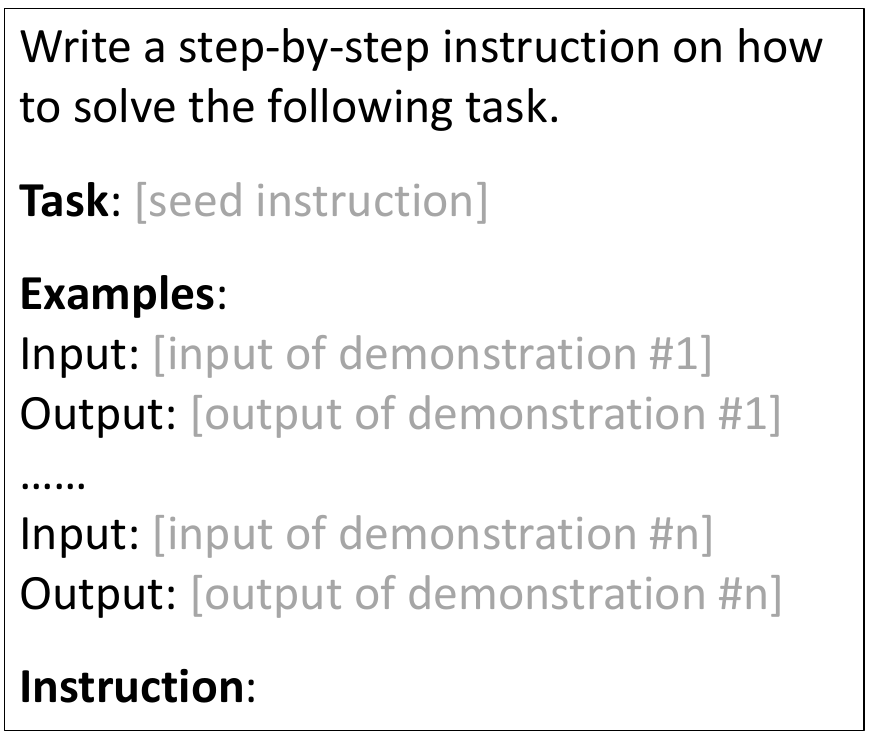}
    \caption{Example explanation instruction}
    \label{fig:explain-example}
  \end{subfigure}
  % \hfill
  \begin{subfigure}[b]{0.48\textwidth}
    \centering
    \includegraphics[width=0.9\textwidth]{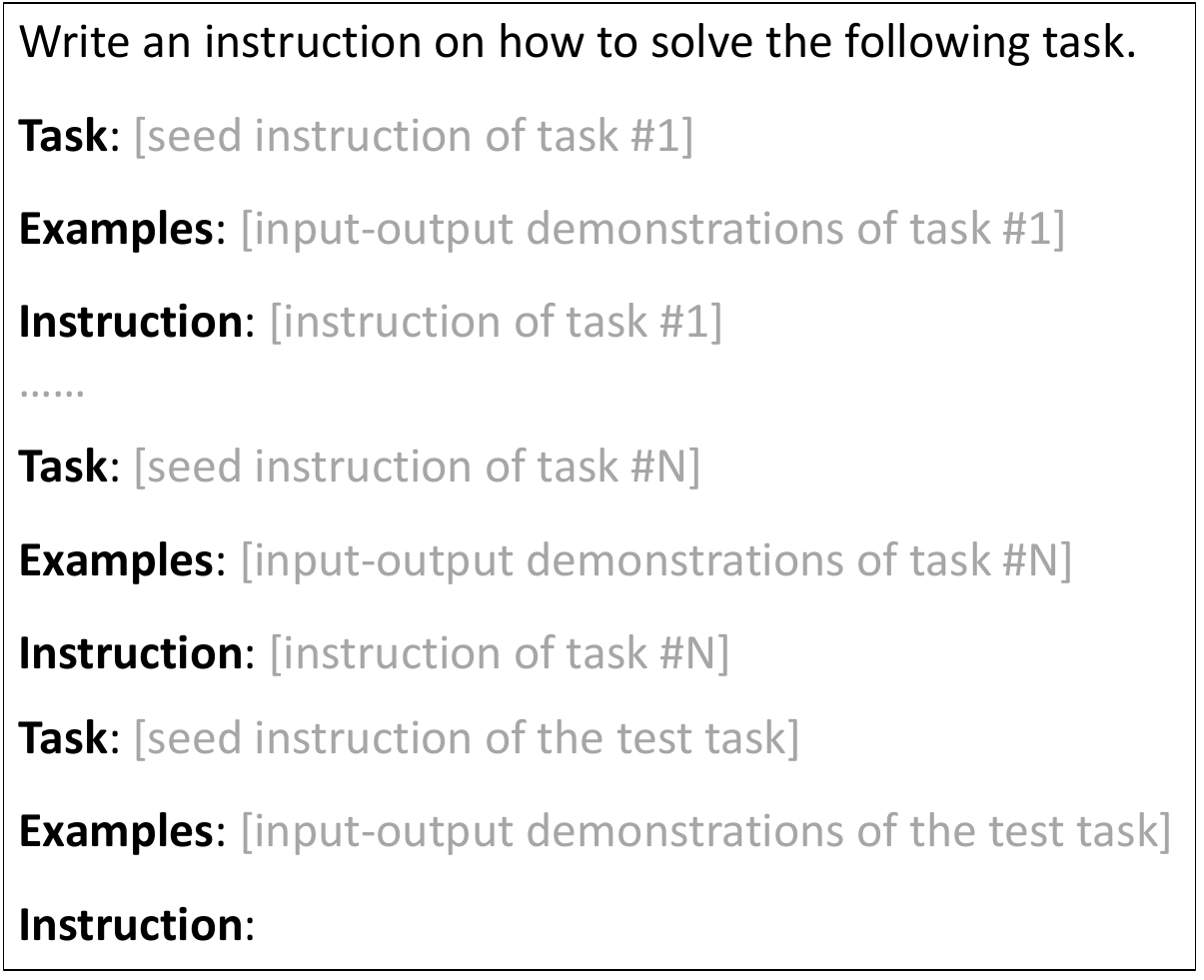}
    \caption{Use other tasks as demonstrations}
    \label{fig:demo-task}
  \end{subfigure}
  \caption{Meta-prompts that we use to specify different desired styles of instructions during instruction generation. For Figure~\ref{fig:demo-task}, we collect 3 groups of demonstration tasks with varying average instruction length (short, medium, long), so as to guide the LLM to generate instructions of different granularities.}
  \label{app_fig:all-meta-prompts}
\end{figure*}

In this section, we list all meta-prompts utilized during instruction generation, as outlined in \S\ref{sec:generation}. For the zero-shot setting, we omit the ``Examples'' field in the meta-prompt to let the LLM rephrase the seed instruction. Besides, the meta-prompt with explanations to the demonstrations is not applicable in the zero-shot setting. The meta-prompt that uses other tasks as demonstrations (Figure~\ref{fig:demo-task}) is integrated with three groups of demonstration tasks, each varying in the average instruction length. Therefore, the LLM is prompted to generate instructions of similar granularity to the demonstration tasks. Demonstration tasks are sampled from SuperNI. In SuperNI, each task is paired with a concise \textit{task summary} and a detailed \textit{task definition} which is usually much longer. For each demonstration task, we use the task summary as the seed instruction and the task definition as the target instruction. We abstain from utilizing the task definition in test tasks because (1) some task definitions are too long to fit in the T5 model together with the input (2) we practically find that the LLM tends to repeat the task definition to a large extent if it is used as the seed instruction. Although Auto-Instruct has never seen the much longer task definition of test tasks, our selected instruction still performs better than using the task definition as the instruction, which holds an average score of 62.41 on SuperNI in the few-shot setting. We leave the exploration of integrating more complicated instructions as future work.

\begin{figure*}[t]
    \centering
    \includegraphics[width=1.0\textwidth]{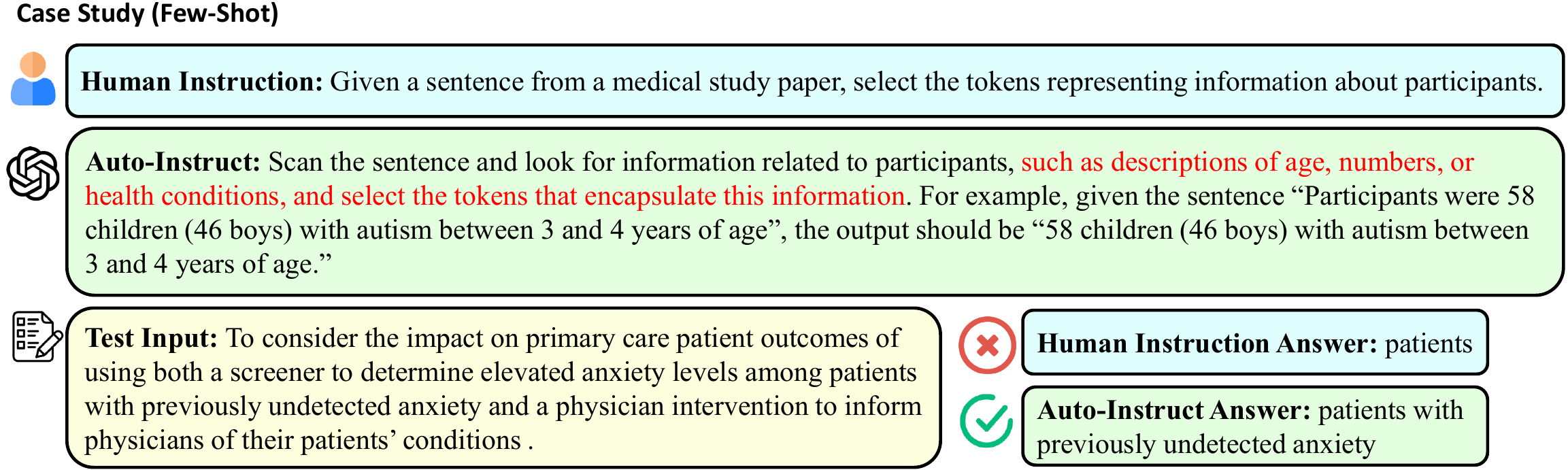}
    \caption{A case study in the few-shot setting, where the few-shot demonstrations are omitted for brevity. The initial human-written instruction provides general guidelines for extracting participant information from a medical study. However, it does not specify the scope of such information, leading to an answer that only includes the keyword “patients” but ignores other relevant information. In contrast, Auto-Instruct provides an instruction that delineates the types of participant information to be extracted (highlight in {\color{red}red}), after seeing the output formats of the demonstrations. Prompted by the improved instruction which suggests health conditions are examples of the requested information, the LLM generates a more comprehensive answer that incorporates the patients’ health conditions, \textit{i.e.}, “with undetected anxiety”.}
    \label{app_fig:case_1}
\end{figure*}

\begin{figure*}[t]
    \centering
    \includegraphics[width=1.0\textwidth]{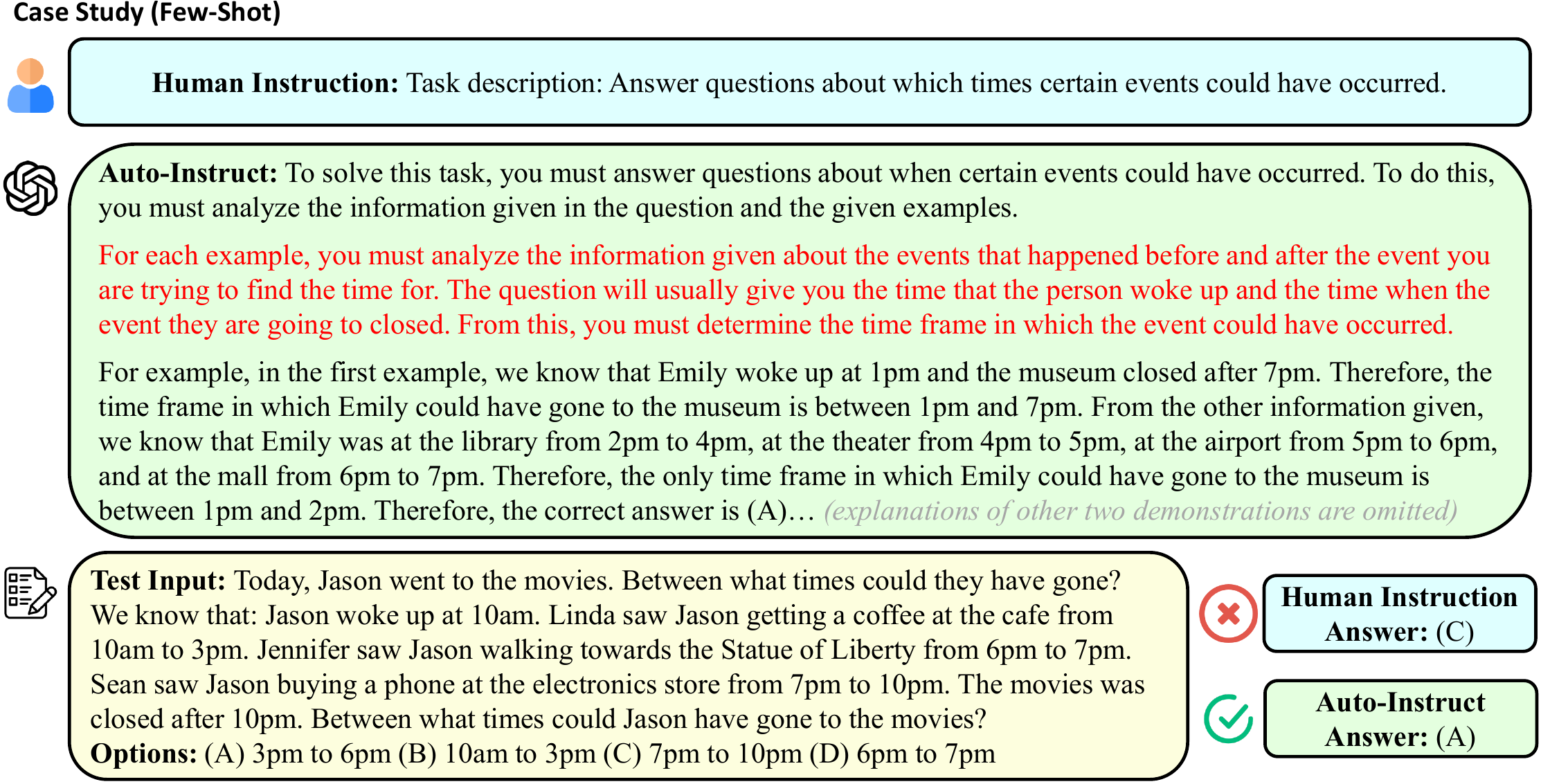}
    \caption{A case study in the few-shot setting, where the few-shot demonstrations are omitted for brevity. The human instruction provides a general and concise description of the question’s requirements. In contrast, generated based on the demonstrations, the instruction from Auto-Instruct offers a more concrete description about the information present in the input and emphasizes which aspects should be focused on (highlight in {\color{red}red}). Besides, Auto-Instruct provides explanations of the few-shot demonstrations as complementary information for the LLM to understand these examples.
}
    \label{app_fig:case_2}
\end{figure*}

\begin{figure*}[t]
    \centering
    \includegraphics[width=1.0\textwidth]{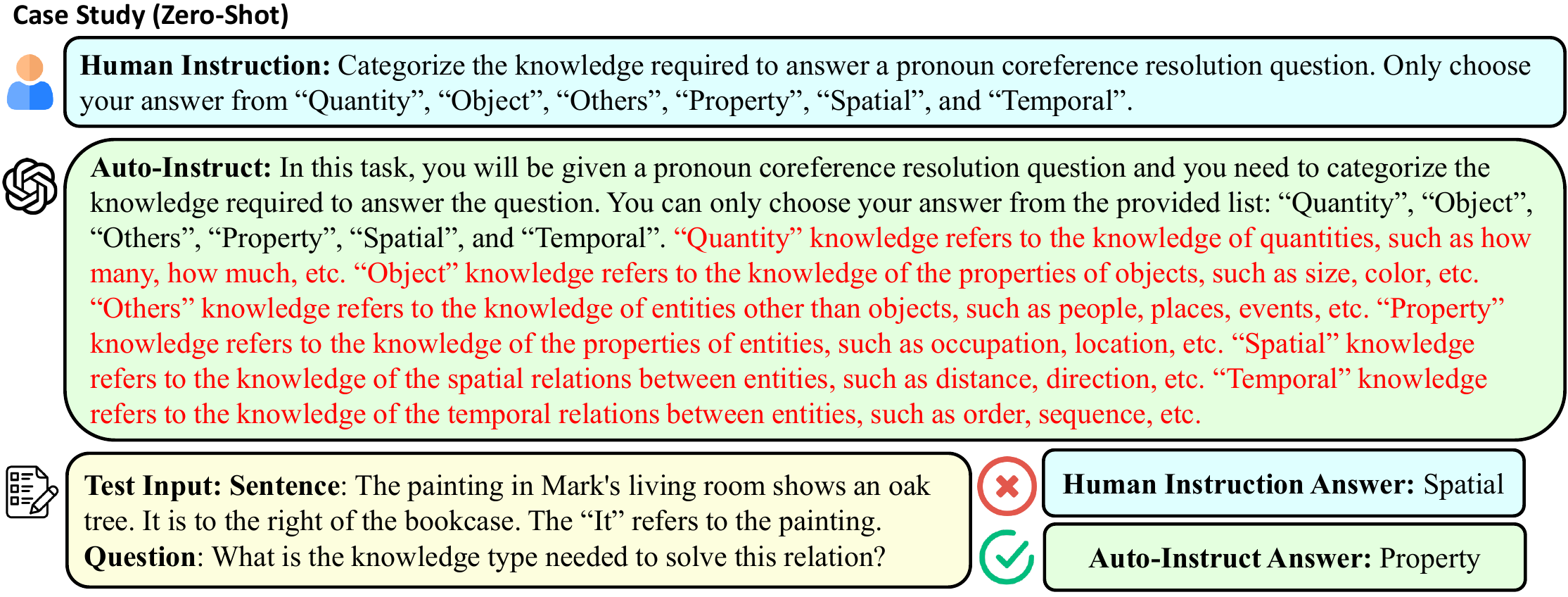}
    \caption{In this zero-shot classification case, the human-written instruction only provides the name of each category. As a result, the LLM can only attempt to determine the target category based on these single-word surface names, which often lack sufficient clarity for differentiation. In contrast, the instruction provided by Auto-Instruct explains the meaning of each category, which greatly facilitates the LLM’s comprehension of these categories. While Auto-Instruct tends to over-interpret when explaining the “Others” category, most of the additional information (highlight in {\color{red}red}) are useful for making more accurate predictions.}
    \label{app_fig:case_3}
\end{figure*}

\section{Additional Case Study}
\label{app:case_study}
In this section, we provide 3 more cases (2 few-shot and 1 zero-shot) where Auto-Instruct improves the original human-written instructions. These case studies are shown in Figure~\ref{app_fig:case_1},~\ref{app_fig:case_2}, and~\ref{app_fig:case_3}. Please refer to the corresponding captions for detailed case explanations.

\begin{table*}[t]
\centering
\resizebox{1.0\textwidth}{!}{
\begin{tabular}{l|ll}
\toprule
\textbf{Task Category} & \multicolumn{2}{c}{\textbf{Task Names}} \\
\midrule
\multirow{3}{*}{\makecell{Coherence\\Classification}} & task066\_timetravel\_binary\_consistency\_classification & task070\_abductivenli\_incorrect\_classification \\
 & task1573\_samsum\_classification & task065\_timetravel\_consistent\_sentence\_classification \\
 & task298\_storycloze\_correct\_end\_classification & \\
\midrule
\multirow{3}{*}{Data to Text} & task1728\_web\_nlg\_data\_to\_text & task1407\_dart\_question\_generation \\
 & task677\_ollie\_sentence\_answer\_generation & task1409\_dart\_text\_generation \\
 & task1598\_nyc\_long\_text\_generation & task957\_e2e\_nlg\_text\_generation\_generate \\
\midrule
\multirow{5}{*}{\makecell{Answerability \\Classification}} & task349\_squad2.0\_answerable\_unanswerable\_question\_classification & task226\_english\_language\_answer\_relevance\_classification \\
 & task020\_mctaco\_span\_based\_question & task290\_tellmewhy\_question\_answerability \\
 & task1439\_doqa\_cooking\_isanswerable & task1442\_doqa\_movies\_isanswerable \\
 & task242\_tweetqa\_classification & task1624\_disfl\_qa\_question\_yesno\_classification \\
 & task520\_aquamuse\_answer\_given\_in\_passage & task050\_multirc\_answerability \\
\midrule
\multirow{11}{*}{\makecell{Information\\Extraction}} & task1506\_celebrity\_minimal\_dob\_span & task1517\_limit\_classfication \\
 & task456\_matres\_intention\_classification & task388\_torque\_token\_classification \\
 & task1518\_limit\_answer\_generation & task1410\_dart\_relationship\_extraction \\
 & task676\_ollie\_relationship\_answer\_generation & task180\_intervention\_extraction \\
 & task749\_glucose\_reverse\_cause\_emotion\_detection & task684\_online\_privacy\_policy\_text\_information\_type\_generation \\
 & task958\_e2e\_nlg\_text\_generation\_parse & task1413\_dart\_object\_identification \\
 & task292\_storycommonsense\_character\_text\_generation & task578\_curiosity\_dialogs\_answer\_generation \\
 & task1597\_nyc\_slot\_filling & task747\_glucose\_cause\_emotion\_detection \\
 & task678\_ollie\_actual\_relationship\_answer\_generation & task1510\_evalution\_relation\_extraction \\
 & task1451\_drug\_dose\_extraction & task683\_online\_privacy\_policy\_text\_purpose\_answer\_generation \\
 & task179\_participant\_extraction & task1411\_dart\_subject\_identification \\
 & task181\_outcome\_extraction & task748\_glucose\_reverse\_cause\_event\_detection \\
 & task621\_ohsumed\_yes\_no\_numerical\_answer\_generation & task647\_answer\_generation \\
\midrule
\multirow{12}{*}{\makecell{Commonsense\\Classification}} & task1210\_atomic\_classification\_madeupof & task1215\_atomic\_classification\_capableof \\
 & task1216\_atomic\_classification\_causes & task1202\_atomic\_classification\_xneed \\
 & task136\_winowhy\_knowledge\_categorization & task1196\_atomic\_classification\_oeffect \\
 & task291\_semeval\_2020\_task4\_commonsense\_validation & task1208\_atomic\_classification\_xreason \\
 & task1206\_atomic\_classification\_isbefore & task1197\_atomic\_classification\_oreact \\
 & task1213\_atomic\_classification\_desires & task116\_com2sense\_commonsense\_reasoning \\
 & task1201\_atomic\_classification\_xintent & task1198\_atomic\_classification\_owant \\
 & task1212\_atomic\_classification\_hasproperty & task1203\_atomic\_classification\_xreact \\
 & task1214\_atomic\_classification\_xwant & task1200\_atomic\_classification\_xeffect \\
 & task1209\_atomic\_classification\_objectuse & task1204\_atomic\_classification\_hinderedby \\
 & task1207\_atomic\_classification\_atlocation & task1205\_atomic\_classification\_isafter \\
 & task1199\_atomic\_classification\_xattr & \\
\midrule
\multirow{4}{*}{Word Analogy} & task1156\_bard\_analogical\_reasoning\_tools & task1159\_bard\_analogical\_reasoning\_containers \\
 & task1155\_bard\_analogical\_reasoning\_trash\_or\_treasure & task1157\_bard\_analogical\_reasoning\_rooms\_for\_containers \\
 & task1154\_bard\_analogical\_reasoning\_travel & task1158\_bard\_analogical\_reasoning\_manipulating\_items \\
 & task1152\_bard\_analogical\_reasoning\_causation & task1153\_bard\_analogical\_reasoning\_affordance \\
\midrule
\multirow{2}{*}{Code to Text} & task131\_scan\_long\_text\_generation\_action\_command\_long & task129\_scan\_long\_text\_generation\_action\_command\_short \\
 & task110\_logic2text\_sentence\_generation & \\
\midrule
\multirow{5}{*}{\makecell{Dialogue\\Generation}} & task1603\_smcalflow\_sentence\_generation & task1714\_convai3\_sentence\_generation \\
 & task360\_spolin\_yesand\_response\_generation & task574\_air\_dialogue\_sentence\_generation \\
 & task565\_circa\_answer\_generation & task576\_curiosity\_dialogs\_answer\_generation \\
 & task1600\_smcalflow\_sentence\_generation & task1729\_personachat\_generate\_next \\
 & task1730\_personachat\_choose\_next & task361\_spolin\_yesand\_prompt\_response\_classification \\
\bottomrule
\end{tabular}}
\caption{All SuperNI test tasks, grouped into different categories. These task categories are not seen during the training of the instruction ranking model. Besides, any task that is sourced from the same original dataset as any test task is excluded from training.}
\label{app_tab:test_tasks}
\end{table*}

\section{All Test Tasks}
\label{app:test_tasks}
In Table~\ref{app_tab:test_tasks}, we list all 91 SuperNI test tasks used in our out-of-domain experiments. Since the size of tasks is imbalanced on SuperNI, for efficient evaluation, we randomly sample 200 instances for each task, making a total of 18,200 test examples.

% This is a section in the appendix.

\end{document}